\documentclass[logo, copyright, nonumbering]{style}

\usepackage[utf8]{inputenc} %
\usepackage[T1]{fontenc}    %
\usepackage{hyperref}       %
\usepackage{url}            %
\usepackage{booktabs}       %
\usepackage{amsfonts}       %
\usepackage{nicefrac}       %
\usepackage{microtype}      %
\usepackage{xcolor}         %
\usepackage{xspace}

\usepackage{graphicx}
\usepackage{bm}
\usepackage{amsmath}
\usepackage{amssymb}
\usepackage{todonotes}
\usepackage{pifont}
\usepackage{multirow}
\usepackage{multicol}
\usepackage{wrapfig}
\usepackage{colortbl}
\usepackage{caption}
\usepackage{bbm}
\usepackage{tabularx}
\usepackage{etoolbox}
 \usepackage[frozencache=true,cache=true]{minted}
\usepackage[most,minted,listings]{tcolorbox}

\newtcolorbox{abstractboxstyle}[1][]{
    enhanced,         
    breakable,       
    colback=citecolor!3, 
    colframe=black!15,
    boxrule=0.0pt,      
    left=12pt,            
    right=12pt,          
    bottom=10pt,        
    boxsep=2pt, 
    before skip=\medskipamount, 
    after skip=\medskipamount,  
    title=Abstract,                 
    fonttitle=\Large\bfseries\centering, 
    coltitle=black,                 
    colbacktitle=citecolor!3,
    toptitle=9pt,                   
    bottomtitle=5pt,                
    before upper={\vspace{2pt}}, 
    top=4pt, 
    #1 
}

\RenewEnviron{abstract}{%
    \begin{abstractboxstyle}
      \normalsize
      \BODY  
    \end{abstractboxstyle}
}

\newcolumntype{Y}{>{\raggedright\arraybackslash}X}
\title{Collision- and Reachability-Aware Multi-Robot Control with Grounded LLM Planners}
\runningtitle{Constraint Grounded Multi-Robot Planner}

\newcommand{\aspace}{\hspace{1em}}
\newcommand{\ucsb}{$^{1}$}
\newcommand{\Mit}{$^{2}$}

\newcommand{\ibm}{$^{4}$}
\newcommand{\cisco}{$^{5}$}

\author[*]{
    \textbf{Jiabao Ji}\ucsb \thanks{Correspondance: <\url{jiabaoji@ucsb.edu}>, <\url{chang87@ucsb.edu}>} 
    \aspace 
    \textbf{Yongchao Chen}$^{2,3}$
    \aspace 
    \textbf{Yang Zhang}\ibm  \\
    \textbf{Ramana Rao Kompella}\cisco \aspace 
    \textbf{Chuchu Fan}\Mit \aspace
    \textbf{Gaowen Liu}\cisco \aspace
    \textbf{Shiyu Chang}\ucsb
}
\affil[1]{UC Santa Barbara}
\affil[2]{MIT}
\affil[3]{Harvard University}
\affil[4]{MIT-IBM Watson AI Lab}
\affil[5]{Cisco Research}

\begin{document}

\newcommand{\fullplan}{\textsc{FullPlan}}
\newcommand{\fullplansp}{\textsc{FullPlan}~}
\newcommand{\replan}{\textsc{Replan}}
\newcommand{\replansp}{\textsc{Replan}~}
\newcommand{\boxnet}{\texttt{BoxNet}}
\newcommand{\boxnetsp}{\texttt{BoxNet}~}
\newcommand{\boxnettwod}{\texttt{BoxNet2D}}
\newcommand{\boxnettwodsp}{\texttt{BoxNet2D}~}
\newcommand{\boxnetthreed}{\texttt{BoxNet3D}}
\newcommand{\boxnetthreedsp}{\texttt{BoxNet3D}~}

\newcommand{\hlg}[2]{\setlength{\fboxsep}{0.3pt}\colorbox{green!#2}{\rule[-.05\baselineskip]{0pt}{.7\baselineskip}{#1}}}
\newcommand{\hlr}[2]{\setlength{\fboxsep}{0.3pt}\colorbox{red!#2}{\rule[-.05\baselineskip]{0pt}{.7\baselineskip}{#1}}}
\newcommand{\hlb}[2]{\setlength{\fboxsep}{0.3pt}\colorbox{aqua!#2}{\rule[-.05\baselineskip]{0pt}{.7\baselineskip}{#1}}}
\definecolor{grey}{rgb}{0.8,0.8,0.8}
\definecolor{aqua}{rgb}{0, 1, 1}
\definecolor{steel}{rgb}{0.2734, 0.5078, 0.7031}
\definecolor{slate}{rgb}{0.1836, 0.3086, 0.3086}
\definecolor{tableheadcolor}{RGB}{200,200,200}
\definecolor{transgray}{gray}{0.9}
\definecolor{lightblue}{rgb}{0.18,0.39,0.62}
\definecolor{blue2}{rgb}{0.1,0.5,0.65}
\definecolor{green2}{RGB}{0, 100, 0}
\definecolor{pink}{rgb}{0.8,0.4,0.4}
\definecolor{darkred}{RGB}{165, 42, 42}
\definecolor{lightbluebg}{rgb}{0.96,0.98,1.0}
\definecolor{rulegray}{rgb}{0.7,0.7,0.8}

\definecolor{orange2}{RGB}{221, 132, 82}
\definecolor{red2}{RGB}{196, 78, 82}
\definecolor{purple2}{RGB}{149, 108, 180}
\definecolor{darkblue2}{RGB}{76, 114, 176}
\definecolor{lightblue2}{RGB}{100, 181, 205}
\newcommand{\myrowcolour}{\rowcolor{tableheadcolor}}

\newtcblisting{fancyminted}[4][]{%
  listing only,
  listing engine=minted,
  before={\refstepcounter{listing}\label{#4}},
  breakable,
  minted language=#2,
  minted style=colorful,
  minted options={
    breaklines,
    fontsize=\scriptsize,
    numbersep=5pt,
    escapeinside=||,
    gobble=0
  },
  fontupper=\ttfamily,
  colback=lightbluebg,
  colframe=rulegray,
  coltitle=black,
  colbacktitle=white,
  coltitle=black,
  boxrule=0.3mm,
  arc=2mm,
  outer arc=2mm,
  left=6mm,
  right=6mm,
  top=3mm,
  bottom=3mm,
  enhanced,
  attach boxed title to top left={yshift=-2mm, xshift=4mm},
  boxed title style={
    colframe=rulegray,
    colback=white,
    sharp corners,
    boxrule=0.3mm
  },
  title=\textbf{Listing~\number\numexpr\value{listing}+1\relax} $\vert$ {#3}, 
  #1
}
\newcommand{\website}{\url{https://github.com/UCSB-NLP-Chang/constraint-grounded-planner}\xspace}
\maketitle

\begin{abstract}
Large language models (LLMs) have demonstrated strong performance in various robot control tasks. However, their deployment in real-world applications remains constrained. Even state-of-the-art LLMs, such as GPT-o4mini, frequently produce invalid action plans that violate physical constraints, such as directing a robot to an unreachable location or causing collisions between robots. 
This issue primarily arises from a lack of awareness of these physical constraints during the reasoning process.
To address this issue, we propose a novel framework that integrates reinforcement learning with verifiable rewards (RLVR) to incentivize knowledge of physical constraints into LLMs to induce constraints-aware reasoning during plan generation. In this approach, only valid action plans that successfully complete a control task receive positive rewards. 
We applied our method to two small-scale LLMs: a non-reasoning Qwen2.5-3B-Instruct and a reasoning Qwen3-4B.
The experiment results demonstrate that constraint-aware small LLMs largely outperform large-scale models without constraints, grounded on both the \boxnetsp task and a newly developed \boxnetthreedsp environment built using MuJoCo. 
This work highlights the effectiveness of grounding even small LLMs with physical constraints to enable scalable and efficient multi-robot control in complex, physically constrained environments. 
\iconbar{
    \href{https://github.com/UCSB-NLP-Chang/constraint-grounded-planner}{\faGithub~GitHub}\quad
    \href{https://huggingface.co/collections/Shiyu-Lab/constraint-grounded-multi-robot-llm-planner-6834dde02542148d1901d963}{\huggingfaceicon{}~HuggingFace}
}
\end{abstract}

\section{Introduction}

Robotic control task requires controllers to find action plans given the robot's physical constraints. 
Conventional methods often employ planning tools, such as PDDL~\cite{fox2003pddl2} and temporal logics~\cite{emerson1990temporal} to find optimal plans. However, they often demand expert knowledge to convert task constraints to formal language and struggle to scale efficiently in multi-robot systems due to increased search time~\cite{chen2024autotamp, chen2025code0as0symbolic0planner0, huang2022inner}. 
Recent advances in Large Language Models (LLMs), which excel at complex reasoning tasks like math and coding~\cite{deepcoder2025, deepseekai2025deepseekr1, shao2024deepseekmath, code-r1}, have inspired their application in robotic control. 
LLMs can interpret natural-language task instructions and generate valid action plans~\cite{meng2025audere0, chen2024scalable, chu2025rlgeneralize}; for instance, ChatGPT can effectively generate high-level commands such as \textit{``Robot A, move the square object to panel 2''}~\cite{chen2024scalable, mandi2023roco}. Paired with low-level execution functions that translate these commands into control signals for robots, they have proven successful in various multi-robot tasks~\cite{chen2024scalable, mandi2023roco, sun2022multi}.

However, these successes have mainly been in synthetic or constrained environments, where physical interactions are overly simplified. For example, most tasks in RocoBench have predefined all the possible valid robot interactions with the objects, largely restricting the action space for LLMs. 
This has led to significant issues in real-world scenarios, where LLM planners tend to violate many basic physical constraints. 
In particular, two important constraints are often overlooked. 

$\bullet$ \textbf{Reachability constraint:} LLM would direct a robot arm to an unreachable position~\cite{chen2024scalable, zhang2025gcbf+}. 

$\bullet$ \textbf{Collision constraint:} LLM would schedule robots to the same space, leading to collisions~\cite{mandi2023roco, jones2025beyond}. 

\begin{figure}[t]
    \centering
    \includegraphics[width=\linewidth]{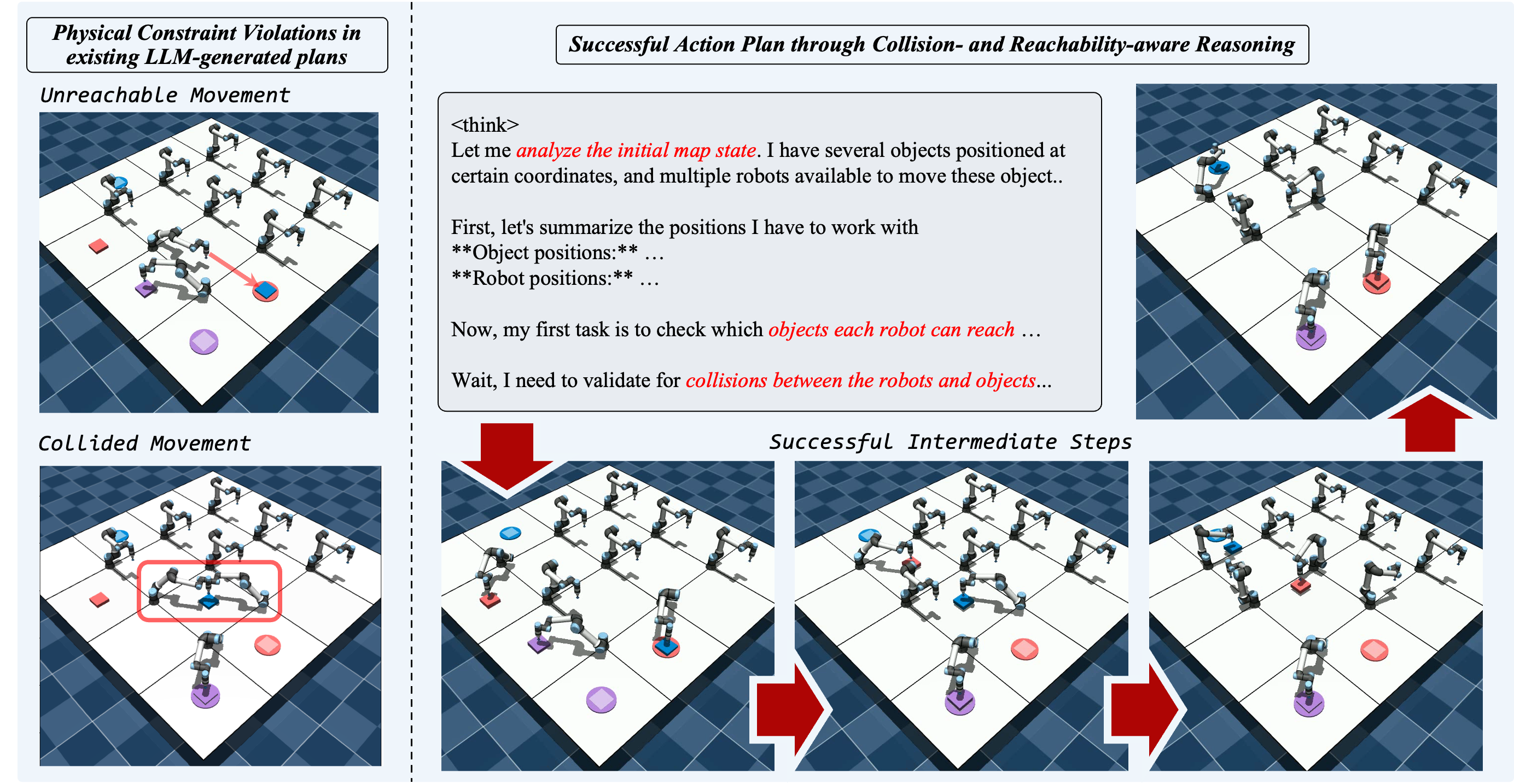}
    \caption{Illustration of LLM-based multi-robot control. (Left) Without grounding constraint knowledge, the LLM generates action plans that result in unreachable positions or collisions. (Right) Our planner generates valid movement actions through constraint-aware reasoning (highlighted in \textcolor{red}{red}) that successfully completes the \boxnetsp task after grounding robotic constraints knowledge.}
    \label{fig:intro-plan-comparison}
\end{figure}

As an example, Figure~\ref{fig:intro-plan-comparison} (left) shows invalid actions generated by a SOTA reasoning LLM GPT-o4mini, which easily violate these constraints, leading to significant safety and feasibility concerns.

These issues highlight the imperative to equip LLM planners with the ability to understand, analyze, and adhere to basic physical constraints. 
However, incorporating these constraints would require strong geometric reasoning and self-reflection capabilities, particularly when the number of robots is large, which may pose nontrivial challenges to LLMs. 
This raises a key research question: \textit{Can LLMs, given their current reasoning capabilities, be trained to integrate physical constraints into the planning process? If so, to what extent can they succeed?}

To study these research questions, this paper presents a novel framework to incentivize these physical constraints into LLM planners, enabling them to reason about action validity during plan generation.
Specifically, we leverage reinforcement learning with verifiable rewards (RLVR) that incorporates checks for reachability, kinematic feasibility, and collision avoidance. By using binary success/failure signals derived from the robot control environment, we ensure that the LLM only receives rewards for generating physically valid plans. 
This fine-tuning process enables the LLM to reason about the validity during plan generation, leading to more reliable and collision-free action plans. 

Our experiments on two LLMs, a non-reasoning Qwen2.5-3B-Instruct and a reasoning Qwen3-4B, have shown several encouraging findings. 
First, by incorporating the physical constraints into the reward, LLM planners can quickly acquire the ability to adhere to the physical constraints, thus drastically increasing the planning success rate, outperforming SOTA large-scale LLMs. For example, our best planner can achieve 0.87 and 0.53 pass@1 on two \boxnet-task multi-robot datasets, while the best baseline planner can only achieve 0.37 and 0.33 pass@1, respectively.
Figure~\ref{fig:intro-plan-comparison} (right) shows the thinking process and the generated plan by our fine-tuned LLM, which successfully solves the task without violating physical constraints.
Second, our reasoning probing experiments have revealed that LLMs indeed learn to correctly identify whether the geometric constraints are satisfied or not. Finally, such capabilities acquired from RL can generalize to unseen geometric configurations, which further verifies that LLMs learn the generic geometric reasoning skills rather than overfitting to specific geometric configurations.

In summary, the contributions of this work are as follows:
\begin{itemize}
    \item We propose a novel framework that grounds LLMs with knowledge of action validity and collision constraints, ensuring LLM-planner-generated plans avoid unreachable positions, object collisions, or robot collisions.
    \item We introduce two new environments based on \boxnetsp task, which incorporate realistic physical constraints and serve as testbeds for evaluating LLM-based multi-robot control.
    \item We implement our approach on two small-scale LLMs, demonstrating that even small models like Qwen2.5-3B-Instruct and Qwen3-4B—when grounded with physical constraints—can outperform larger, state-of-the-art LLMs in complex multi-robot control tasks.
\end{itemize}
\section{Method}

\subsection{Overview}
In this section, we introduce our framework for grounding LLMs with reachability and collision awareness. Denote $\mathcal{M}_{\bm \theta_0}(\cdot)$ as the initial LLM for performing the robot control tasks,  which is capable of generating an action plan  $\bm{s} \sim \mathcal{M}_{\bm \theta_0}(\bm q; \mathcal{C})$ for solving the given control task described by $\bm q$ under a set of physical constraints $\mathcal{C}$, such as the reachability of a robot arm and collision avoidance. 
Our goal is to fine-tune the LLM such that the generated solution $\bm s$ successfully moves objects to their target positions while not violating the constraints $\mathcal{C}$. 
In the following, we first introduce the RLVR framework for grounding physical constraints in Section~\ref{subsec:method-rlvr}, then introduce our initial LLM policy warmup strategy in Section~\ref{subsec:method-warmup}, and two different planner modes we consider in Section~\ref{subsec:method-planmode}.

\subsection{Grounding LLM with Physical Constraints through RLVR}\label{subsec:method-rlvr}
We adopt a similar RL framework to the DeepSeek-R1 LLM~\cite{deepseekai2025deepseekr1, ren2025deepseek0prover0v20}, which employs the group relative policy optimization (GRPO)  algorithm~\cite{shao2024deepseekmath}. 
Specifically, at each training step $i, i \geq 1$, we sample a group of different plans and its corresponding reasoning $\{\bm{s}_1, \bm{s}_2, \dots, \bm{s}_G\}$ from the old LLM policy $\mathcal{M}_{\bm{\theta}_{i-1}}$ for each query robot-control task $\bm q$, where $G$ is the group size. Each plan $\bm s_j$ is simulated in a manually implemented environment with physical constraints. The corresponding reward function $r(\cdot)$ later estimates whether it successfully completes the given task with the simulation environment feedback. Then the LLM is optimized by maximizing the following objective~\cite{shao2024deepseekmath}. 
\begin{equation*}
\begin{split}
    \mathcal{J}_{GRPO}(\mathcal{M}_{\bm \theta_i}) &= \mathbb{E} \left[ \left( \bm q \sim \mathcal{D}, \{ \bm s_j \}_{j=1}^G \sim \mathcal{M}_{\bm \theta_{i-1}}(\bm O|\bm q; \mathcal{C}) \right) \right]  \\
    & = \frac{1}{G} \sum_{j=1}^G \left( \min \left( \frac{\mathcal{M}_\theta(\bm s_j | \bm q; \mathcal{C}))}{\mathcal{M}_{\bm \theta_{i-1}}(\bm s_j | \bm q; \mathcal{C}))} A_j, \text{clip} \left( \frac{\mathcal{M}_\theta( \bm s_j | \bm q; \mathcal{C}))}{\mathcal{M}_{\bm \theta_{i-1}}( \bm s_j | \bm q; \mathcal{C}))}, 1 - \epsilon, 1 + \epsilon \right) A_j \right) \right. \\
    & \quad \left. - \beta \mathbb{D}_{KL}\left( \mathcal{M}_{\bm \theta_{i}} \big\| \mathcal{M}_{\bm \theta_0} \right) \right) ,
\end{split}
\label{eq:grpo-obj}
\end{equation*}
where $\mathcal{D}$ denotes the training data and $A_j$ represents the advantage, computed as the reward of each plan subtracted by the average reward within the group. Detailed definitions are in Appendix~\ref{sec:appendix-implemenation}.

Our reward function, denoted as $r(\cdot)$ largely follows the design in DeepSeek-R1~\cite{deepseekai2025deepseekr1}, with an additional plan efficiency term. Specifically,
\begin{equation*}
\begin{split}
r(\bm s; \bm q, \bm{s}^*, \mathcal{C}) = r_{\text{format}}(\bm s) + r_{\text{execute}}(\bm s; \mathcal{C}) - r_{\text{efficiency}}(\bm s; \bm s^*),
\end{split}
\label{eq:reward}
\end{equation*}
$r_{\text{format}}(\bm s)=0.1$ if the generated solution adheres to the required thinking-then-response format and $0$ otherwise.  $r_{\text{execute}}(\bm s; \mathcal{C}) = 1$ if the simulator verifies that the plan \ding{182} accomplishes the task AND \ding{183} no physical constraints are violated, and $0$ otherwise. Incorporating physical constraint checking in $r_{\text{execute}}$ is the key mechanism to improve constraint awareness of the LLM planner. Finally,
\[
r_{\text{efficiency}}(\bm{s}; \bm{s}^*) = \max\left(0,\ 0.1 \times \left( \text{len}(\bm{s}) - \text{len}(\bm{s}^*) \right) \right)\]
penalizes the excessive length compared to a golden plan, $\bm s^*$, which is obtained by an A* search algorithm (see Appendix~\ref{sec:appendix-implemenation} for search details).
Additionally, we implement a minimum cap of $2 \times r_{\text{format}} (\bm s)$ when $\bm s$ is a valid plan that successfully solves the task to ensure that correct plans always receive a higher reward than incorrect ones.

\begin{table}[t]
\small
\centering
\caption{Example synthsized reasoning trace for \fullplansp planner and \replansp planner. We highlight three reasoning behaviors: 1. environment analysis in \textcolor{blue2}{blue}, 2. validity verification in \textcolor{green2}{green}, and 3. efficiency consideration in \textcolor{orange}{orange}. The environment feedback is marked in \textcolor{gray}{gray}.
}
\label{tab:example-trace}
\vspace*{2mm}
\begin{tabularx}{\textwidth}{Y|Y}
\toprule \midrule
\textbf{\fullplansp Planner Reasoning}: & \textbf{\replansp Planner Reasoning}: \\ 

\begin{minipage}[t]{\linewidth}
\textcolor{gray}{Object positions:} \\
\textcolor{gray}{\hspace*{2em} Object 0: [0.75, 1.75] $\dots$ }\\
\textcolor{blue2}{ <think> Okay, let me analyze the given environment before coming up with a movement plan $\dots$.} \\
\textcolor{green2}{\#\#\# Evaluating Reachability:} \\
\textcolor{green2}{- **Robot 0** cannot reach either object $\dots$ } \\
\textcolor{green2}{\#\#\# Collision Checks:} \\
\textcolor{green2}{- No objects are placed at the same position$\dots$} \\
\textcolor{orange}{Let me finalize this, since there are no collisions, this plan efficiently ensures all objects are moved $\dots$} 
\end{minipage}
& 

\begin{minipage}[t]{\linewidth}
\textcolor{gray}{<observation>} \\
\textcolor{gray}{Object positions:} \\
\textcolor{gray}{\hspace*{2em} Object 0: [0.75, 2.75] $\dots$} \\
\textcolor{gray}{</observation> }\\
\textcolor{blue2}{<think> Alright, let's break down the situation $\dots$ }\\
\textcolor{gray}{<observation>} \\
\textcolor{gray}{Object positions:} \\
\textcolor{gray}{{\hspace*{2em}  Object 1: [2.25, 2.75] $\dots$}}\\
\textcolor{gray}{</observation>} \\
\textcolor{blue2}{<think> Alright, let's analyze this $\dots$ }
\end{minipage}
\\

\midrule
\bottomrule
\end{tabularx}

\end{table}

\begin{figure}[t]
    \centering
    \includegraphics[width=0.97\linewidth]{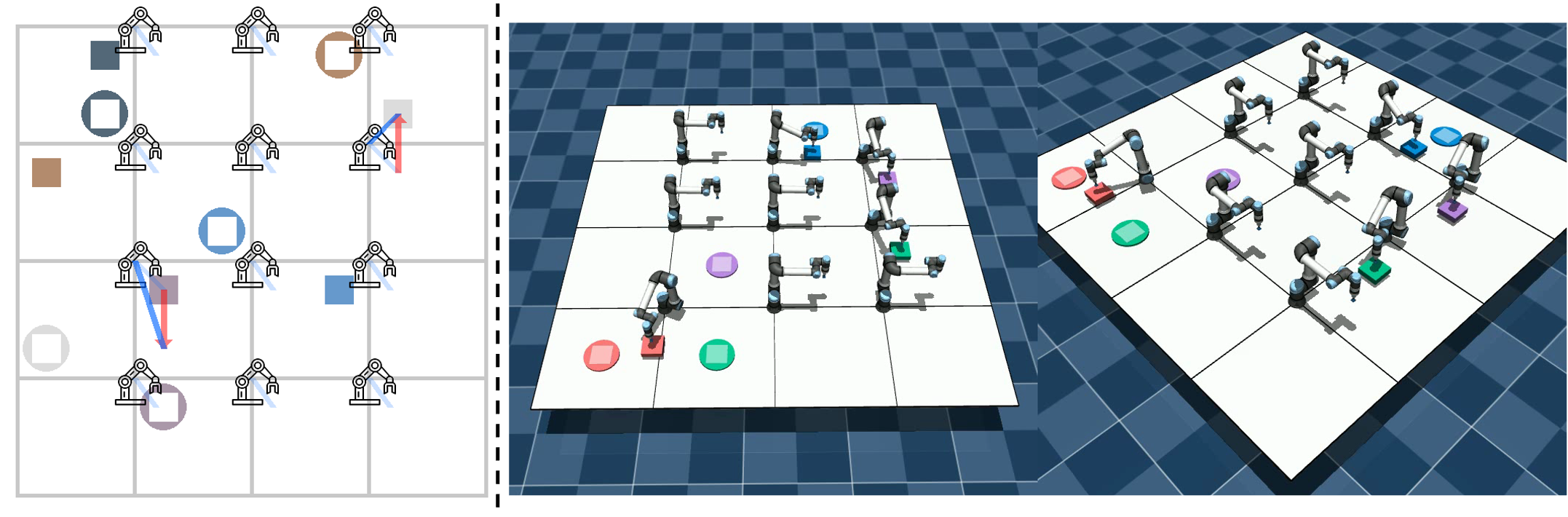}
    \caption{(Left) An example \boxnettwodsp environment. The blue lines mark the robot arm, and the red lines mark the movement trajectory. (Right) An example \boxnetthreedsp environment. Both environments require robots to collaborate to move boxes to the circle with the corresponding color.}
    \label{fig:boxnet-example}
\vspace*{-0.15in}
\end{figure}

\subsection{Initial Supervised Fine-Tuning (SFT) Warmup}
\label{subsec:method-warmup}
Prior works have shown that LLMs' initial performance on a reasoning task is crucial to RLVR training~\cite{deepcoder2025, shao2024deepseekmath, code-r1}. Since off-the-shelf LLMs often struggle with robotic control tasks, we introduce an SFT warmup to equip them with basic robot control knowledge before RL training.

The SFT data need to contain two components: \ding{182} a correct plan to solve a given task, and \ding{183} a reasoning chain that reflects a multi-step decision-making process leading to the correct plan. To synthesize such data, for each task, we first use the A* search algorithm to search for the optimal plan. Then, we pass the plan to an LLM, which is prompted to generate a reasoning process for the plan, consisting of the following three patterns:

$\bullet$ \textit{Analysis of the given environment}, where the LLM assesses the current positions of robot arms and objects, \textit{e.g.}, \textit{let me analyze the current situation};

$\bullet$ \textit{Validity verification}, where the LLM reasons about an arm's reachable area based on its base position and potential collision between different arms, \textit{e.g.}, \textit{If Robot 0 moves, $\dots$, it will collide};

$\bullet$ \textit{Efficiency considerations}, where the LLM evaluates whether multiple movements can be parallelized to improve the plan efficiency.

Table~\ref{tab:example-trace} shows the example reasoning chain synthesized by GPT4o-mini, where the three patterns are rendered in different colors. Appendix~\ref{sec:appendix-prompts} shows the full prompt for our reasoning synthesis.

\subsection{Two Planners: \fullplansp Planner and \replansp Planner}\label{subsec:method-planmode}
We consider two different LLM-based planners in this work. The first planner, referred to as \fullplan, involves the LLM directly generating the entire plan that may take multiple execution steps for solving a task based on the initial positions of all objects and robot poses in the environment. 
The second planner, denoted as \replan, generates one step at a time, observing the updated object positions from the environment (appended to its context) before generating the next step.
This allows the planner to evolve dynamically as the environment changes through multiple execution steps.
Table~\ref{tab:example-trace} provides the example planning processes for two different LLM-based planners for the same initial environment. We highlight that the \replansp planner has access to multiple intermediate observations of the environment, while the \fullplansp planner only sees the initial environment.

\section{\boxnet-Based Multi-Robot Environments}\label{sec:environment-description}

In this work, we primarily experiment with \boxnetsp task~\cite{chen2024scalable}, where multiple robots collaborate to move objects across different cells to targeted locations in a fixed grid map. 
This section details two environments we developed, a modified \boxnettwodsp environment and the newly developed \boxnetthreedsp environment, both equipped with realistic physical constraint checks.

\paragraph{\boxnettwod.}
Figure~\ref{fig:boxnet-example} (left) shows a \boxnettwodsp environment. In this environment, robot arms are placed at a corner of a grid environment. Each arm can reach its neighboring grids for picking and placing objects.
Unlike the previous \boxnetsp environment that predefined all valid robot arm actions, we allow LLMs to generate spatial coordinates directly, significantly expanding the action space. For example, the action \textit{``Robot0 Move (1.25, 1.25) $\rightarrow$ (1.75, 1.75), False''} moves \textit{Robot0}'s arm to \textit{(1.75, 1.75)} without picking up an object. In contrast, \textit{``Robot1 Move (2.25, 1.75) $\rightarrow$ (1.25, 1.25), True}'' indicates \textit{Robot1} picking up the object at \textit{(1.75, 1.25)} and moving it to \textit{(2.25, 1.75)}.

We pre-define four points within each grid, \textit{e.g.}, \textit{(0.25, 0.25), (0.25, 0.75), (0.75, 0.25), (0.75,0.75)}, for object placement and robot arm moving, and later we will show that the fine-tuned LLM can generalize to other points in experiments.
Three physical constraints are implemented: 
\ding{182} \textit{reachability verification}, which checks whether the target position of a robot is unreachable.
\ding{182} \textit{robot collision detection}, which checks whether the movement trajectory of different arms intersects with each other, or one robot's movement trajectory intersects with another robot arm, leading to a collision.
\ding{183} \textit{object collision detection}, which checks whether two objects are placed at the same spatial coordinates during the plan execution.
Example invalid actions of \boxnettwodsp are provided in Appendix~\ref{sec:appendix-example}.

\paragraph{\boxnetthreed.}
Figure~\ref{fig:boxnet-example} (right) shows a \boxnetthreedsp environment. In this environment, we employ the UR5e robot arm as the basic robot arm\footnote{\url{https://www.universal-robots.com/products/ur5e/}}. Similar to the 2D environment,  the goal is to move colored boxes into corresponding circles of the same color with the fewest actions. 
Each robot arm's base is fixed and moves its arm around to reach different grids for object picking and placement. We employ an RRT planner implemented by RoCoBench~\cite{mandi2023roco} for low-level control signal generation given LLM-generated coordinates for arm position movement\footnote{The RRT planner is adapted from the implementation in RoCoBench official code base (\url{https://github.com/MandiZhao/robot-collab/blob/main/rocobench/rrt.py})}.
We employ Mujoco as the simulation engine~\cite{todorov2012mujoco}, which provides the arm reachability check and collision detection mechanism. 

\section{Experiment}

\begin{table*}[t]
\vspace*{-0.15in}
\centering
{
\caption{{Performance of different LLM planners on \boxnettwodsp and \boxnetthreedsp. For each model, we report the results for \fullplansp planner and \replansp planner side-by-side (\fullplansp / \replan). }} 
\label{tab:main-exp}
\small
\begin{tabular}{l|ccc|ccc}
\toprule \midrule
\multirow{2}{*}{\hspace*{2em} Model} & \multicolumn{3}{c|}{\boxnettwod} & \multicolumn{3}{c}{\boxnetthreed} \\
 & \small{Success $\uparrow$} & \small{StepDiff. $\downarrow$} & \small{Para. $\uparrow$} & \small{Success $\uparrow$} & \small{StepDiff. $\downarrow$} & \small{Para. $\uparrow$} \\
\midrule
\rowcolor{transgray}
\multicolumn{7}{c}{{Search Algorithm}} \\ \midrule
A* & 1 & 0 & 2.24 & 1 & 0 & 2.14 \\
\midrule
\rowcolor{transgray}
\multicolumn{7}{c}{{LLMs without constraint knowledge grounding}} \\ \midrule
\small{GPT-4omini}       & 0.06 / 0.05 & 2.35 / 0.14  & 1.17 / 1.15 & 0.07 / 0.06 & 0.79 / 0.45 & 1.03 / 1.08 \\
\small{GPT-4o}           & 0.12 / 0.11 & 2.14 / 0.13  & 1.15 / 1.22 & 0.10 / 0.12 & 0.23 / 0.68 & 1.35 / 1.11 \\
\small{GPT-o4mini}       & 0.37 / 0.35 & 0.24 / -0.31 & 1.58 / 1.87 & 0.11 / 0.33 & 0.14 / 1.21 & \textbf{1.45} / 1.53 \\
\small{Qwen2.5-3B-Inst}  & 0.0 / 0.0 & ----- / ----- & ----- / ----- & 0.08 / 0.0 & 0.20 / ----- & 1.40 / ----- \\
\small{Qwen2.5-7B-Inst}  & 0.02 / 0.02 & 1.45 / 0.31  & 1.20 / 1.23 & 0.05 / 0.08 & 0.41 / 0.35 & 1.13 / 1.07 \\
\small{QwQ-32B}          & 0.04 / 0.07 & 0.35 / 0.17  & 1.12 / 1.21 & 0.07 / 0.15 & 0.24 / -0.09 & 1.08 / 1.31 \\
\small{Qwen3-4B}         & 0.14 / 0.13 & 0.23 / 0.14  & 1.29 / 1.29 & 0.15 / 0.11 & 0.07 / 0.31  & 1.17 / 1.14 \\
\small{Qwen3-8B}         & 0.18 / 0.15 & -0.23 / -0.34 & 1.24 / 1.31 & 0.17 / 0.13 & -0.02 / 0.09 & 1.22 / 1.21 \\
\small{Qwen3-14B}        & 0.19 / 0.21 & -0.31 / -0.24 & 1.34 / 1.41 & 0.10 / 0.14 & 0.17 / 1.37 & 1.34 / 1.35 \\
\small{Qwen3-32B}        & 0.11 / 0.14 & 0.17 / -0.03 & 1.24 / 1.12 & 0.14 / 0.17 & 0.09 / 0.04 & 1.27 / 1.09 \\
\midrule 
\rowcolor{transgray}
\multicolumn{7}{c}{{LLMs with grounded constraint knowledge}} \\ \midrule
\small{Qwen2.5-3B-SFT}   & 0.34 / 0.30 & 0.11 / -0.04 & 1.51 / 1.39 & 0.27 / 0.39 & -0.07 / -0.05 & 1.27 / 1.39 \\
\small{Qwen2.5-3B-RL}    & 0.58 / 0.68 & -0.65 / 0.23 & 1.53 / 1.50 & 0.42 / 0.48 & -0.15 / -0.14 & 1.33 / 1.49 \\
\small{Qwen3-4B-SFT}     & 0.45 / 0.31 & -0.12 / -0.15 & \textbf{1.92} / 1.35 & 0.37 / 0.43 & 0.09 / -0.11 & 1.32 / 1.48 \\
\small{Qwen3-4B-RL}      & \textbf{0.87} / \textbf{0.75} & \textbf{-0.84} / \textbf{-0.64} & 1.73 / \textbf{1.64} & \textbf{0.45} / \textbf{0.53} & \textbf{-0.25} / \textbf{-0.29} & 1.39 / \textbf{1.56} \\
\midrule
\bottomrule
\end{tabular}
}
\vspace*{-0.1in}
\end{table*}

In this section, we conduct empirical experiments on the two \boxnet-based environments to assess the effectiveness of our method. We first present the experiment setup in Section~\ref{sec:experiment-setup} and then the experiment results in Section~\ref{sec:experiment-result}, followed by additional ablation studies in Section~\ref{sec:ablation}.

\subsection{Experiment Setup}\label{sec:experiment-setup}

\paragraph{Dataset generation.}
We create environments with various map sizes and object initial and target positions for both \boxnettwodsp and \boxnetthreedsp.
Specifically, for \boxnettwodsp, we use map sizes ranging from $2\times 2$ to $6\times 6$ and 1 to 5 objects, resulting in 55,000 training and 250 testing environments. For \boxnetthreedsp, we use map sizes from $2\times 2$ to $4\times4$ with 1 to 3 objects, yielding 1,800 training and 160 testing environments. 
The object position is randomly sampled from the pre-defined points, while the robot arms are evenly placed at the grid joints to ensure that all grids in the map can be reached.
For each randomly sampled environment, the manually implemented A* search algorithm verifies that a valid solution action plan exists.
Detailed dataset statistics are summarized in Appendix~\ref{subsec:appendix-boxnet}. 

\paragraph{Evaluation metric.}
We evaluate LLM-based planners mainly from two perspectives: \ding{182} \textit{Success}, the proportion of generated plans that solve given robotic tasks, measured by \textit{pass@1} over four trials per environment; and \ding{183} \textit{StepDiff.}, the difference in number of steps between successful plans and the best plan among A* solutions. We also report \textit{Para.}, the maximum number of robots operating in parallel in any intermediate step of a successful plan.

\paragraph{Baseline LLMs.}
We mainly compare with off-the-shelf LLMs. To ensure comprehensive coverage of existing LLMs, our evaluation includes both reasoning and non-reasoning models, closed-source and open-source ones across different parameter scales. Specifically, we consider closed-source LLMs, GPT-4o, GPT-4o-mini, and GPT-o4-mini. On the open-source side, we include Qwen-2.5 and Qwen3 series, with parameter sizes ranging from 3B to 32B.

\paragraph{Training details.}
We use two base LLMs: a non-reasoning LLM Qwen-2.5-3B-Instruct and a reasoning LLM Qwen3-4B. For SFT warm-up, we use a learning rate of $1e-5$ for Qwen-2.5-3B-Instruct and $3e-5$ for Qwen-3-4B with the AdamW optimizer~\cite{loshchilov2017adamw}. Training runs for 10 epochs on \fullplansp and 5 epochs on \replansp.
RLVR training uses a fixed $1e-6$ learning rate for 200 steps with the GRPO algorithm~\cite{guo2025}. Batch sizes are 256 (group size 8) for \boxnettwodsp and 64 for \boxnetthreedsp. Following prior work~\cite{deepcoder2025, code-r1}, we set $\beta=0$ in the GRPO loss. We use the VeRL framework~\cite{sheng2024verl}, and run all experiments on $2\times8$ NVIDIA H100 GPUs.

\subsection{Experimental Results}\label{sec:experiment-result}
\begin{wrapfigure}{r}{0.4\textwidth}
\vspace*{-0.22in}
    \centering
    \includegraphics[width=0.95\linewidth]{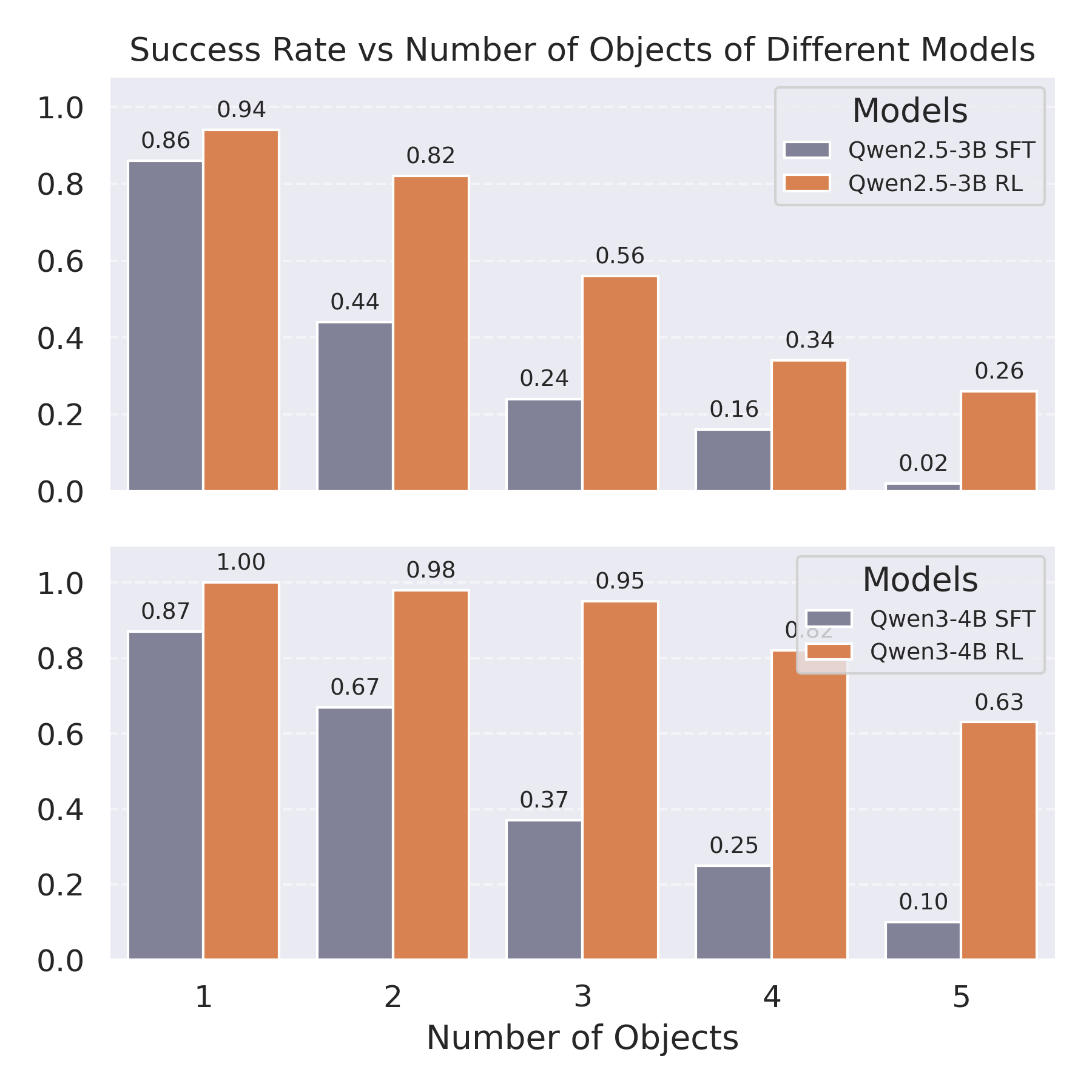}
    \caption{Success rate against number of boxes in the \boxnettwodsp environment.}
\label{fig:success-boxnumber}
\vspace*{-0.15in}
\end{wrapfigure}
\paragraph{Grounding empowers small-scale LLMs to outperform larger ones.}
We first evaluate the grounded LLM planner performance in Table~\ref{tab:main-exp}.
The LLMs with physical constraints knowledge grounded through SFT warmup and further RL training are denoted by the suffix {\textit{-SFT}} and {\textit{-RL}}, respectively.
We highlight three observations:
First, grounding constraint knowledge significantly boosts planning performance,
enabling 3B and 4B LLMs to achieve higher success against larger ones.
For example, Qwen3-4B-RL \fullplansp planner achieves 0.87 success rate, 0.5 higher than the best baseline GPT-o4mini. 
Second, grounded LLM planners produce more efficient plans than the A* search algorithm on solved tasks. For example, Qwen3-4B-RL has 0.84 fewer steps than the ground-truth plan from our A* implementation, showing a strong reasoning ability and also echoes the findings in prior works that compare LLM planners with A* on Sudoku~\cite{lehnert2024beyondastar, su2024dualformer}.
Third, planner performance on \boxnetthreedsp is generally worse than on \boxnettwod. This suggests that, although we applied multiple feasibility checks in \boxnettwod, some physical constraints remain missing. The \boxnetthreedsp environment uses a more advanced simulation engine and thus exposes more limitations.
These results underscore the importance of developing realistic robotic environments that capture real-world complexity for future LLM-based robotic control research.

Figure~\ref{fig:success-boxnumber} visualizes planner performance against numbers of boxes for \boxnettwod.
We highlight that RL-trained planners better preserve performance when task complexity increases.  For example, the performance gap between Qwen3-4B-SFT and Qwen3-4B-RL grows from 0.13 to 0.53 when the number of boxes increases from 1 to 5 for \boxnettwod, highlighting better scalability of RL planners.

\begin{wrapfigure}{r}{0.6\textwidth}
\vspace*{-0.15in}
\centering
\captionof{table}{Planning performance generalization on unseen \boxnettwodsp environments. }
\label{tab:transfer-exp}
\resizebox{\linewidth}{!}{
\begin{tabular}{l|cc|cc}
\toprule \midrule 
\multirow{2}{*}{\hspace*{2em}{Model}}  & \multicolumn{2}{c|}{\textsc{RandRob}} & \multicolumn{2}{c}{\textsc{NewCoord}} \\
& \small{Success $\uparrow$} & \small{StepDiff. $\downarrow$} & \small{Success $\uparrow$} & \small{StepDiff. $\downarrow$}  \\ \midrule
\rowcolor{transgray}
\multicolumn{5}{c}{\fullplansp Planner} \\ \midrule
    \small{Qwen2.5-3B-SFT}   & 0.39 & 0.12  & 0.32 & 0.21 \\ 
    \small{Qwen2.5-3B-RL }   & 0.58 & -0.04 & 0.55 & -0.32 \\ 
    \small{Qwen3-4B-SFT  }   & 0.48 & 0.03  & 0.43 & -0.03 \\  
    \small{Qwen3-4B-RL   }   & \textbf{0.79} & \textbf{-0.40}  & \textbf{0.87} & \textbf{-0.39} \\  \midrule
\rowcolor{transgray}
\multicolumn{5}{c}{\replansp Planner} \\ \midrule
    \small{Qwen2.5-3B-SFT}   & 0.39 & 0.23  & 0.33 & \textbf{0.09} \\ 
    \small{Qwen2.5-3B-RL }   & 0.71 & 1.24  & 0.68 & 1.04 \\ 
    \small{Qwen3-4B-SFT  }   & 0.41 & -0.03 & 0.37& 0.15\\  
    \small{Qwen3-4B-RL   }   & \textbf{0.75} & \textbf{-0.15} & \textbf{0.69} & \textbf{0.09} \\  
\midrule
\bottomrule
\end{tabular}
}
\end{wrapfigure}

\paragraph{RL planners generalize better to unseen environments.}
To measure how the planners' reasoning ability generalizes, we evaluate the planners' performance in two unseen variants of \boxnettwodsp test data:
\ding{182} Random robot layout, denoted as \textsc{RandRob}, where the robot positions are randomly assigned on the grid joints in maps ranging from $2\times 2$ to $5\times5$. We ensure that all testing data are solvable, which means every box can reach its target position via robot movement.
\ding{183} Unseen coordinates, \textit{i.e.}, \textsc{NewCoord}), where the initial and target position coordinates of all objects in \boxnettwodsp test set are perturbed by a random offset $(\Delta x, \Delta y) \sim \mathcal{U}([-0.2, 0.2]^2)$.
Example data are shown in Appendix~\ref{sec:appendix-example}.

Table~\ref{tab:transfer-exp} reports the performance of our grounded planners on two unseen environments. 
We highlight that the RL-trained planners consistently outperform SFT ones while maintaining plan efficiency.
For example, Qwen3-4B-RL \fullplansp planner achieves 0.87 success rate on \textsc{NewCoord}, 0.44 better than the SFT variant, showing better generalization of the reasoning capability. This observation also aligns with previous RL for LLM works~\cite{chu2025rlgeneralize, razin2025makes}.

\paragraph{Reasoning behavior change after RL.}
Previous results have shown that RL training significantly improves the planner's planning ability. In this section, we analyze in more detail how the reasoning behavior of the LLM-based planners changes before and after RL fine-tuning.

\begin{wrapfigure}{r}{0.6\textwidth}
\vspace*{-0.15in}
\centering
\captionof{table}{Number of different reasoning behaviors for grounded \fullplansp planners.}
\label{tab:reason-behavior}
\resizebox{0.93\linewidth}{!}{
\begin{tabular}{l|cc|cc|cc}
\toprule \midrule 
\multirow{2}{*}{\hspace*{2em} Model} & \multicolumn{2}{c|}{\boxnettwod} & \multicolumn{2}{c|}{\textsc{RandRob}}  & \multicolumn{2}{c}{\textsc{NewCoord}} \\ 
~ & \small{Rea.} & \small{Col.} & \small{Rea.} & \small{Col.} & \small{Rea.} & \small{Col.} \\ \midrule
Qwen2.5-3B-SFT & 8.0   & 9.8   & 8.4 & 8.3 & 8.9 & 9.1 \\ 
Qwen2.5-3B-RL & 8.6   & 10.4  & 9.2 & 8.9 & 9.8 & 9.5 \\  \midrule
Qwen3-4B-SFT & 9.1   & 7.4   & 7.3 & 6.4 & 10.1 & 9.4\\ 
Qwen3-4B-RL & 10.1  & 8.2   & 7.7 & 6.7 & 10.3 & 9.6 \\ \midrule
\bottomrule
\end{tabular}
}
\end{wrapfigure}

Given the critical role of reachability checks and collision checks in generating successful action plans, we prompt GPT-4o to count the number of these checks in the reasoning traces produced by our \fullplansp planners across three \boxnettwodsp environment variants. The prompts used are provided in Appendix~\ref{sec:appendix-prompts}. 
As shown in Table~\ref{tab:reason-behavior}, the RL planners perform more reachability checks (Rea.) and collision checks (Col.) than the initial SFT planner. These checks help ensure the feasibility of action plans and lead to a large improvement in success rate. This observation suggests that RL training helps the LLM better understand the importance of these checks and use them more consistently.

\begin{figure}[t]
    \centering
    \includegraphics[width=0.97\linewidth]{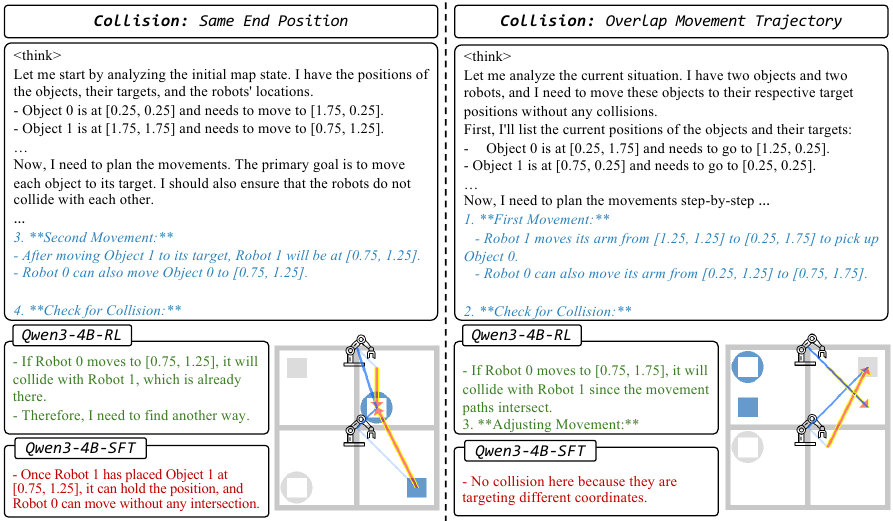}
    \caption{Example reasoning trace generated by grounded \fullplansp planners. Invalid action plans are manually inserted into the trace history and highlighted in \textcolor{blue2}{blue}. The correct continuations that identify and correct these errors are shown in \textcolor{green2}{green}, while incorrect continuations are shown in \textcolor{darkred}{red}. The bottom figure visualizes the collision between two movements. RL planner better detects errors.}
    \label{fig:perturbed-reasoning-trace}
\end{figure}

We also conduct a qualitative analysis to verify their reasoning ability by injecting error steps into the trace. 
Specifically, we insert an invalid action that would lead to collision into the intermediate reasoning steps. To test whether the planner can recognize and correct such errors, we append the phrase \textit{``Collision Check''} to the perturbed trace to trigger verification. Figure~\ref{fig:perturbed-reasoning-trace} shows the full examples, where the injected invalid actions are highlighted in blue. The LLM's continuation is marked in green if it identifies and corrects the error, and in red if it fails. We find that the RL-trained planner successfully finds the error and traces the issue to same target position and overlapped movement paths. This suggests that RL helps build better physical constraints-aware reasoning.

\subsection{Ablation Study}\label{sec:ablation}

In this section, we explore the design choices for our framework on the \boxnettwodsp environment, focusing on:
\ding{182} How does SFT warmup affect final planner performance?
\ding{183} Is the textual thinking necessary for planner performance?
\ding{184} How does the efficiency penalty affect the planner's behavior?

\begin{wrapfigure}{r}{0.45\textwidth}
    \centering
\small
\vspace*{-0.18in}
    \includegraphics[width=1.0\linewidth]{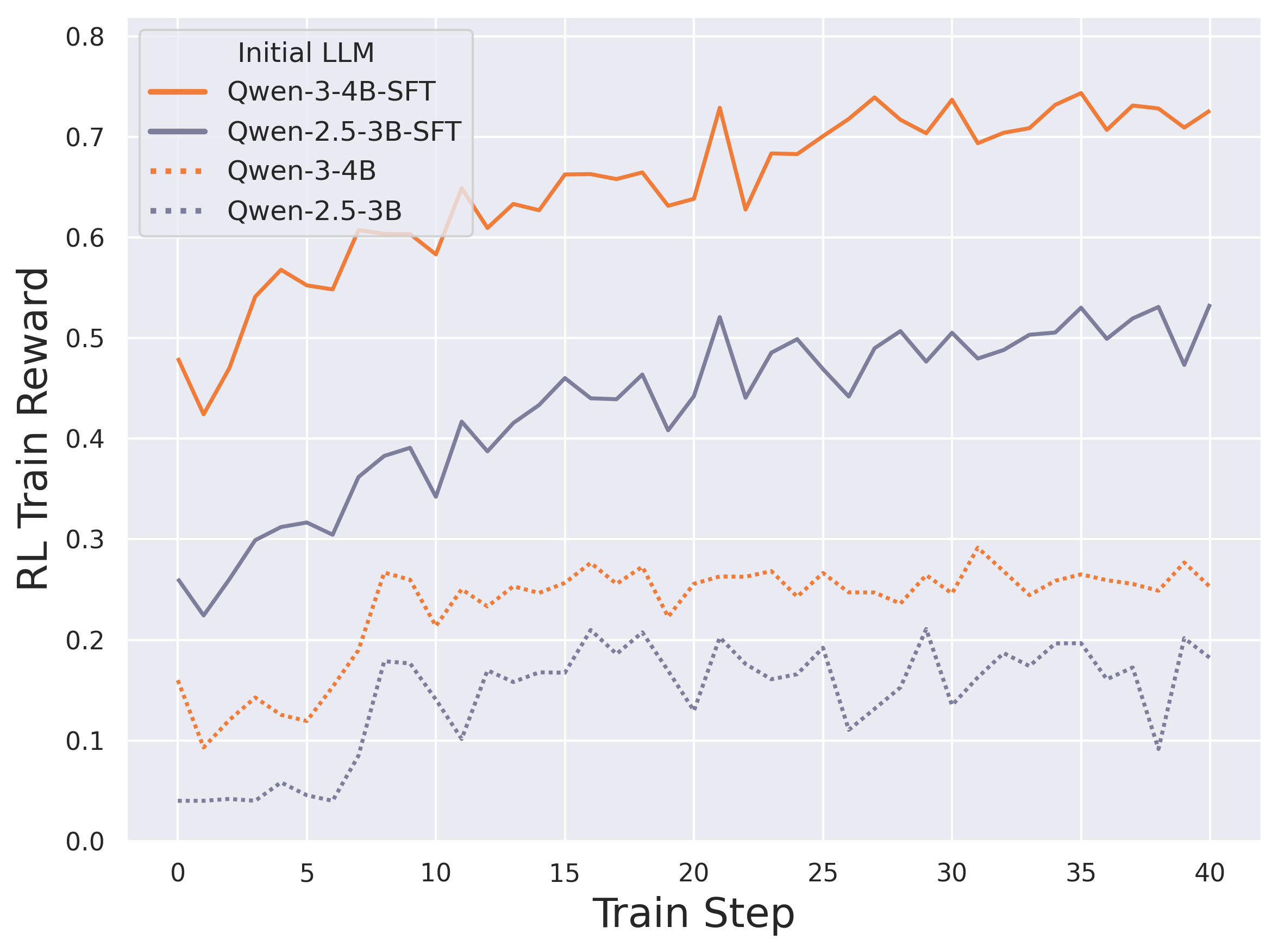}
    \vspace*{-5mm}
    \caption{Training reward trajectory in first 40 steps for different initial LLMs.}
    \label{fig:train-traj}
\vspace*{-0.23in}
\end{wrapfigure}
\paragraph{Initial LLM policy matters.}
Figure~\ref{fig:train-traj} shows how the training reward evolves over the first 40 steps for different initial \fullplansp planners. For the original Qwen2.5-3 B-Instruct and Qwen3-4B models, we observe a sharp reward increase of about 0.1 (the format reward) within the first 10 steps, which indicates that they quickly learn to produce answers in required output format. 
However, after this initial gain, the reward plateaus, indicating limited additional learning to improve planning capability. 
In contrast, the LLM with SFT warmup shows a consistent increasing reward trend. This suggests that the SFT warmup helps build a strong foundation, allowing it to continue learning and optimize effectively in RL training.

\paragraph{Textual reasoning improves planning performance.}
To assess the role of textual reasoning for LLM planners, we perform an ablation study with our SFT and RL pipeline on Qwen-2.5-3B-Instruct. In this experiment, we train a planner without the synthesized thinking steps, \textit{i.e.}, it generates only the final action plan with no textual thinking.
As shown in Table~\ref{tab:2d-ablation}, removing intermediate reasoning leads to a notable performance drop: success rates fall from 0.34 to 0.26 for SFT planner, and from 0.58 to 0.39 for RL planner.
This highlights the importance of textual thinking for LLM planners. 

\begin{wrapfigure}{r}{0.45\textwidth}
\vspace*{-0.15in}
\centering
\captionof{table}{
Impact of ablating thinking and $r_{\text{efficiency}}$ on \boxnettwodsp performance.
}
\label{tab:2d-ablation}
\resizebox{\linewidth}{!}{
\begin{tabular}{l|ccc}
\toprule \midrule 
\multirow{2}{*}{\centering \hspace*{1em} Model} & \multicolumn{3}{c}{\boxnettwod} \\ 
~ & \small{Success $\uparrow$} & \small{StepDiff. $\downarrow$} & \small{Para. $\uparrow$} \\ \midrule
Qwen2.5-SFT & 0.34 & 0.11  & 1.51  \\
\hspace*{1em}$-$ thinking & 0.26  & 0.05 & 1.35\\ \midrule
Qwen2.5-RL & \textbf{0.58} & \textbf{-0.65} & \textbf{1.53} \\
\hspace*{1em}$-$ thinking & 0.39 & -0.07 & 1.43 \\ 
\hspace*{1em}$-$ $r_{\text{efficiency}}$ & 0.52 & 1.44 & 1.07 \\ 
\midrule 
\bottomrule
\end{tabular}
}
\end{wrapfigure}

\paragraph{Efficiency penalty in reward improves plan efficiency.}
We observed a surprising finding that RL-trained LLM planners produce more efficient plans than those generated by our hand-crafted A* search algorithm in Table~\ref{tab:main-exp}, which is likely due to the efficiency penalty term in the reward function.
To further understand its role, we conduct an ablation study on $r_{\text{efficiency}}$. Starting from the same initial LLM policy Qwen-2.5-3B-SFT, we perform RL training without the efficiency penalty for \fullplansp planner. Table~\ref{tab:2d-ablation} presents the results. While both RL-finetuned LLM largely improve the success rate, their plan efficiency differs significantly. The planner trained with $r_{\text{efficiency}}$ produces plans that are 2.09 steps shorter. In contrast, 
the parallelism drops close to 1 without the penalty in RL.
These results underscore the importance of efficiency penalty in reward.

\section{Related Work}

\paragraph{Robotic planning and control with LLMs.} 
Robotic planning and control is a complex task that requires high-level planning under various physical constraints. Traditional approaches typically translate task goals and physical constraints into formal logic specifications, such as Temporal Logic or PDDL~\cite{fox2003pddl2, emerson1990temporal}, and solve them using constraint solvers.
More recently, LLMs have been applied to robotic control due to their strong reasoning capabilities and better scalability compared to constraint solvers. 
For example, some works use LLMs to choose actions from predefined motion primitives~\cite{guan2023leveraging, skreta2023errors, DBLP:conf/iclr/LoulaLDLPG0EFEC25}, while others treat code as an intermediate representation for control~\cite{chen2025code0as0symbolic0planner0,  huang2022inner,meng2025audere0, liang2023code, ahn2022can3,singh2023progprompt}. Hybrid approaches, such as 
AutoTAMP~\cite{chen2024autotamp} and Text2Motion~\cite{lin2023text2motion0}, combine traditional planning tools with LLMs for action planning. 
Another series of works employs multi-LLM discussion for robotic tasks~\cite{chen2024scalable, mandi2023roco, zhang2024towards, guo2024embodied, shen2025enhancing}.
While these methods show promising results, many of them overly simplify physical constraints, limiting their real-world applicability. In this work, we demonstrate that even SOTA LLMs struggle under realistic physical constraints, and further introduce a novel approach that grounds smaller LLMs with this constraint knowledge, which largely improves performance.

\paragraph{Reinforcement learning with verifiable rewards for LLM reasoning.} 
Reinforcement learning (RL) has demonstrated significant promise in enhancing the reasoning capabilities of large language models (LLMs) across a wide range of domains, including mathematics~\cite{ren2025deepseek0prover0v20, guo2025,  zeng2025simplerl0zoo0}, code generation~\cite{code-r1,  deepcoder2025, openai2025competitive}, and complex multi-agent systems~\cite{jin2025search0r10, singh2025agentic, feng2024agile0, openai_operator_2024}. A common paradigm involves training LLMs to optimize for a verifiable reward, such as the correctness of a math solution or whether the generated code passes unit tests, using RL training. 
Many previous works show that the RL training process vastly improves LLM reasoning~\cite{chu2025rlgeneralize,feng2024agile0, pan2025medvlm, shen2025vlm,  Hou2025thinkprune0}. The improvement is often accompanied by emergent reasoning behaviors such as feasibility checks and self-reflection, which are difficult to elicit through supervised fine-tuning alone~\cite{chu2025rlgeneralize, zelikman2024quiet, hosseini2024vstar}. 
In this work, we extend the RLVR to robotic control, with a focus on grounding LLMs with knowledge of physical constraints. Our method leverages RLVR to teach LLMs to reason under the constraints inherent in robotic planning tasks, such as collision avoidance, reachability accordance, and goal satisfaction. By integrating physical verification signals into the training process, the model learns to internalize these constraints as part of its reasoning process. This grounding leads to more robust and reliable control planning decisions in downstream robotic applications.

\section{Conclusion}\label{sec:conclusion}
In this paper, we present a novel framework that grounds LLMs with physical constraint knowledge, such as robot arm reachability and collision avoidance. 
By incorporating these constraints, LLMs are able to reason more effectively about action feasibility and generate efficient and physically viable action plans.
To evaluate our approach, we developed two \boxnet-based multi-robot environments, \boxnettwodsp and \boxnetthreed, both equipped with action feasibility checks. Experiments show that even small-scale LLMs at 3B and 4B parameter size, when grounded with constraint knowledge, significantly outperform larger SOTA LLMs. 
Additional experiments on reasoning behavior and generalization further confirm that our models learn constraint-aware reasoning rather than simply overfitting to training data. 
We discuss limitations and potential societal impacts in Appendix~\ref{sec:appendix-limitation}.

\section{Acknowledgement}
Jiabao Ji and Shiyu Chang acknowledge the support from National Science Foundation (NSF) Grant IIS-2338252, NSF Grant IIS-2207052, Cisco Research Program, and the Accelerate Foundation Models Research Program of Microsoft. The work of Yongchao Chen and Chuchu Fan is partially supported by ONR under Award N00014-22-1-2478 and MIT-IBM Watson AI Lab.

\bibliographystyle{unsrt}
\bibliography{reference}

\appendix

\clearpage
\section{Limitations and Societal Impacts}\label{sec:appendix-limitation}
Our work introduces a novel framework to ground LLMs with physical constraint knowledge for robot control tasks, significantly enhancing their ability to reason about action feasibility during plan generation. This leads to substantial improvements in planning performance. We validate the effectiveness of our approach through experiments in two \boxnet-based multi-robot environments on two small-scale LLMs.
However, this work has two limitations:
\ding{182} Our experiments are limited to the \boxnetsp task due to the high implementation overhead required for other robot control environments. Extending our framework to additional physical constraint-sensitive tasks remains an important direction for future work.
\ding{183} The RL training is conducted at a limited scale due to computational constraints. Unlike typical RL setups in math and coding domains that allow for training over multiple epochs~\cite{code-r1, deepcoder2025}, our training is restricted to just one or two epochs. Despite this limitation, our experimental results already demonstrate strong reasoning capabilities on robotic tasks.

Our work aims to advance the integration of LLMs into robotic control planning, which has many promising societal benefits. By enabling LLMs to better understand and operate within physical constraints, LLMs can help build safer, more reliable, and more efficient multi-robot systems. This can potentially enhance robotic automation domains heavily involving many robots. In particular, improved planning performance can reduce operational errors and increase productivity. However, as with any deployment of AI in real-world decision-making systems, there are potential risks if the planners are deployed with a dangerous purpose. Future extensions of this work should also consider robustness to adversarial scenarios to ensure responsible real-world integration.

\section{Additional Implementation Details}\label{sec:appendix-implemenation}

In this section, we provide more implementation details including: 
the RL training algorithm (Section~\ref{subsec:appendix-grpo}),
the implementation of \boxnettwodsp, \boxnetthreedsp and dataset statistics (Section~\ref{subsec:appendix-boxnet}), and the A* search algorithm for data generation (Section~\ref{subsec:appendix-astar-search}).

\subsection{GRPO Algorithm}\label{subsec:appendix-grpo}
GRPO~\cite{shao2024deepseekmath}, or group relative policy optimization, is a variant of PPO algorithm~\cite{schulman2017proximal} proposed for LLM RL.
In this section, we briefly outline the GRPO algorithm and refer readers to the original paper~\cite{deepseekai2025deepseekr1} for more details.

As we mentioned in the main paper, given an initial LLM policy $\mathcal{M}_{\bm \theta_0}$ and a train dataset $\mathcal D$, the GRPO loss is defined as follows:
\begin{equation*}
\small
\begin{split}
    \mathcal{J}_{GRPO}(\mathcal{M}_{\bm \theta_i}) &= \mathbb{E} \left[ \left( \bm q \sim \mathcal{D}, \{ \bm s_j \}_{j=1}^G \sim \mathcal{M}_{\bm \theta_{i-1}}(\bm O|\bm q; \mathcal{C}) \right) \right]  \\
    & = \frac{1}{G} \sum_{j=1}^G \left( \min \left( \frac{\mathcal{M}_\theta(\bm s_j | \bm q; \mathcal{C}))}{\mathcal{M}_{\bm \theta_{i-1}}(\bm s_j | \bm q; \mathcal{C}))} A_j, \text{clip} \left( \frac{\mathcal{M}_\theta( \bm s_j | \bm q; \mathcal{C}))}{\mathcal{M}_{\bm \theta_{i-1}}( \bm s_j | \bm q; \mathcal{C}))}, 1 - \epsilon, 1 + \epsilon \right) A_j \right) \right. \\
    & \quad \left. - \beta \mathbb{D}_{KL}\left( \mathcal{M}_{\bm \theta_{i}} \big\| \mathcal{M}_{\bm \theta_0} \right) \right) ,
\end{split}
\label{eq:grpo-obj}
\end{equation*}
where $i$ denotes the train step, $G$ denotes the group size, $\bm q$ denotes a textual query describing a robotic control task, $\mathcal{C}$ denotes the constraints in text, $A_{j}$ denotes the advantage for $j$-th rollout $s_j$. The definition of advantage is:
\begin{equation*}
    A_j = \frac{r_j - \text{mean}(\bm r)}{\text{std}(\bm r)},
\end{equation*}
given the reward $\bm r = \{r_1, \dots, r_G\}$ for all LLM rollouts to the query task $\bm q$, following the Generalized Advantage Estimation (GAE)~\cite{schulman2015high}.

\subsection{\boxnetsp Environment Implementation and Statistics}\label{subsec:appendix-boxnet}
We implement two different environments based on the \boxnetsp task, which involves multiple robots in a grid map and collaborating to move objects to the corresponding target positions.
Both environments are implemented in Python. 

\paragraph{\boxnettwod} 
For \boxnettwod, we manually implement the feasibility check by calculating the relative geometric position of robot arms and objects. The map size ranges from $2\times 2$ to $6\times 6$, and the object number ranges from 1 to 5. For each unique map configuration, \textit{i.e.}, a tuple of map width, height, and the object number, we randomly generate at most 150 different object initial and target positions to construct the unique environments in train dataset. 
The testing data consists of the square maps with the width ranging from $2$ to $6$, and the object number ranges from 1 to 5. We generate at most 10 unique environments to construct the test dataset.
The detailed dataset statistics are summarized in Table~\ref{tab:dataset-stats}.

In the unseen environment transfer experiment, we generate two variants of \boxnettwodsp test set: \textsc{RandRob}, where the robot position is not evenly placed at the grid joints, and \textsc{NewCoord}, where the object position coordinates are perturbed with a random offset.
For \textsc{RandRob}, all robots are placed in a connected manner, meaning that all objects can be reached by a robot.

\paragraph{\boxnetthreed}
For \boxnetthreed, we use MuJoCo to implement the feasibility checks such as robot arm collision and object collisions. The map size ranges from $2\times 2$ to $4\times 4$, and the object number ranges from 1 to 4. For each map configuration, we randomly generate at most 100 different environments for the training data and at most 5  for the test data. Detailed dataset statistics are summarized in Table~\ref{tab:dataset-stats}

\begin{table}[t]
    \centering
    \caption{Dataset statistics of \boxnettwodsp and \boxnetthreed. The average steps to complete and parallelism are all based on the optimal plans generated by our manually implemented A* algorithm.}
    \vspace{3mm}
    \begin{tabular}{l|ccc}
    \toprule \midrule 
        Dataset & Sample No. & Avg. Step & Para. \\ \midrule
        \boxnettwod-train & 55000 & 8.13 & 1.73\\
        \boxnettwod-test & 250 & 8.32 & 1.75\\
        \textsc{RandRob} & 200 & 7.06 & 1.49\\ 
        \textsc{NewCoord} & 250 & 8.59 & 1.77 \\ \midrule
        \boxnetthreed-train & 1800 & 6.27 & 1.89 \\
        \boxnetthreed-test & 160 & 5.62 & 1.88 \\
    \midrule \bottomrule
    \end{tabular}
    \label{tab:dataset-stats}
\end{table}

\subsection{A* Search Algorithm}\label{subsec:appendix-astar-search}

We implement an A* search algorithm for solving the generated task. At each search step, the general workflow is: \ding{182} select the current best environment state, \ding{183} generate valid action for a single robot, \ding{184} combine multiple valid actions and check whether they can run in parallel, and \ding{185} update the environment with potential next step actions and put to candidate tools for next search step.

We list a Python reference code below:
\begin{fancyminted}{python}{Reference A* search implementation}{lst:astar}

def astar_search(env: Any, max_iterations: int = 1000): 
    open_set = [] 
    closed_set = set() 

    g_scores: Dict[int, float] = {} 
    came_from: Dict[int, Tuple[Optional[int], Optional[str]]] = {}
    states_cache: Dict[int, EnvStates] = {}

    try:
        initial_state_data = env.get_states()
        current_state = EnvStates(env, current_state_data=initial_state_data)
        initial_hash = current_state.hash()
    except Exception as e:
        raise e 

    g_scores[initial_hash] = 0.0
    came_from[initial_hash] = (None, None) 
    states_cache[initial_hash] = current_state

    heapq.heappush(open_set, (current_state.heuristic(), random.random(), initial_hash))
    
    iterations = 0

    while open_set and iterations < max_iterations:
        iterations += 1
        
        f_val, _, current_hash = heapq.heappop(open_set)

        if current_hash in closed_set:
            continue 

        current_state = states_cache[current_hash] 
        closed_set.add(current_hash) 

        if current_state.is_goal():
            return reconstruct_path(came_from, current_hash)

        potential_next_moves = generate_potential_actions(current_state)
        
        for action_str, next_state_obj in potential_next_moves:
            if next_state_obj is None: 
                continue

            next_hash = next_state_obj.hash()
            if next_hash in closed_set: 
                continue

            cost_of_this_action = 1.0 
            tentative_g_score = g_scores[current_hash] + cost_of_this_action

            if tentative_g_score < g_scores.get(next_hash, float('inf')):
                came_from[next_hash] = (current_hash, action_str)
                g_scores[next_hash] = tentative_g_score
                states_cache[next_hash] = next_state_obj 

                f_score_neighbor = tentative_g_score + next_state_obj.heuristic()
                heapq.heappush(open_set, (f_score_neighbor, random.random(), next_hash))

    return None

class EnvStates:
    _hash_val: Optional[int] = None

    def __init__(self, env: Any, parent_state_data=None, current_state_data=None):
        self.env = env
        self.parent_state_data = parent_state_data
        self.cur_states = current_state_data
        # Assumes env has get_target_pos() and can define retraction height internally or via config
        self.target_positions = self.env.get_target_pos() # Target positions should include Z if relevant

    def hash(self) -> int:
        if self._hash_val is not None:
            return self._hash_val
        self._hash_val = xxhash.xxh64(self.cur_states.tobytes()).intdigest()
        return self._hash_val

    def _box_positions(self):
        self.env.reset(states=self.cur_states)
        return {
            boxname: self.env.get_box_pos(boxname)
            for boxname in self.env.object_names
        }

    def arm_positions(self):
        self.env.reset(states=self.cur_states)
        return {
            robot_name: self.env.get_arm_pos(robot_name)
            for robot_name in self.env.robot_names
        }

    def is_goal(self) -> bool:
        current_box_pos_map = self._box_positions()
        if len(current_box_pos_map) != len(self.target_positions):
            return False

        for box_name, target_val in self.target_positions.items():
            if box_name not in current_box_pos_map:
                return False
            # Use new env method for checking if object is at its target
            if not self.env.is_object_at_target(current_box_pos_map[box_name], target_val, box_name):
                return False
        return True

    def heuristic(self) -> float:
        self.env.reset(states=self.cur_states)
        current_obj_positions_map = self._box_positions()
        
        h = 0.0
        num_matched = 0

        for name, target_pos_val in self.target_positions.items():
            if name in current_obj_positions_map:
                current_pos_val = current_obj_positions_map[name]
                if any(np.isnan(current_pos_val)): # Check for NaN
                    return float("inf")
                # Use env method for calculating distance/cost component for heuristic
                h += self.env.calculate_placement_quality(current_pos_val, target_pos_val, name)
                num_matched +=1
            else: # Object in target not found in current state
                return float("inf") 

        if num_matched != len(self.target_positions): # Not all target objects were found or matched
             return float("inf")
        
        return math.sqrt(h) if h > 0 else 0.0

    def apply_actions(self, action_strings: Union[List[str], str]) -> Optional["EnvStates"]:
        self.env.reset(states=self.cur_states)
        action_input = action_strings
        if isinstance(action_strings, list):
            action_input = "\n".join(action_strings)
            
        out = self.env.simulate_one_step(action_input)
        if out["success"]:
            return EnvStates(
                self.env,
                parent_state_data=self.cur_states,
                current_state_data=self.env.get_states(),
            )
        else:
            return None

# --- Utility Functions (Domain-specific helpers removed, env handles them) ---

def generate_single_robot_action(robot_id: str, state: EnvStates) -> List[str]:
    robot_actions = []
    env = state.env # Get the environment reference
    base_pos = env.get_base_pos(robot_id)
    arm_pos = state.arm_positions()[robot_id]

    for obj_id, obj_pos_val in state._box_positions().items():
        target_pos_val = state.target_positions[obj_id]

        if env.check_reach_range(robot_id, obj_pos_val): # Existing env call for reachability
            if env.is_object_at_target(obj_pos_val, target_pos_val, obj_id):
                continue

            # Get potential next positions for the object from the environment
            potential_next_obj_placements = env.get_valid_next_object_positions(
                obj_id, obj_pos_val, robot_id, base_pos
            )
            current_placement_quality = env.calculate_placement_quality(obj_pos_val, target_pos_val, obj_id)
            
            action_candidates_for_obj = []

            # Try to move the object to a better position
            for next_obj_p in potential_next_obj_placements:
                if env.calculate_placement_quality(next_obj_p, target_pos_val, obj_id) < current_placement_quality:
                    # Check if arm is already at the object
                    if not env.is_arm_at_position(arm_pos, obj_pos_val[:2], robot_id):
                        # Action: Move arm to object
                        action_str = env.format_move_action_string(robot_id, obj_pos_val[:2], False)
                        action_candidates_for_obj.append(action_str)
                    
                    # Action: Move object (arm is now assumed to be at object or will be moved by first action)
                    action_str = env.format_move_action_string(robot_id, next_obj_p[:2], True)
                    action_candidates_for_obj.append(action_str)
            
            # Try to move arm to an alternative/safe position if not productively moving an object
            if not action_candidates_for_obj or all(
                # Check if any action involves carrying (True flag)
                # This logic might need refinement based on how format_move_action_string works
                # or if we have a better way to check if an action is "productive"
                "True" not in act for act in action_candidates_for_obj 
            ):
                alternative_arm_destinations = env.get_alternative_arm_destinations(
                    robot_id, base_pos, arm_pos
                )
                if alternative_arm_destinations:
                    # Environment can decide which one to pick or return just one
                    chosen_alt_dest = alternative_arm_destinations[0] # Take the first one
                    if not env.is_arm_at_position(arm_pos, chosen_alt_dest[:2], robot_id):
                        action_str = env.format_move_action_string(robot_id, chosen_alt_dest[:2], False)
                        action_candidates_for_obj.append(action_str)
            
            for action_str_candidate in action_candidates_for_obj:
                if action_str_candidate not in robot_actions:
                    robot_actions.append(action_str_candidate)
    
    return robot_actions


def verify_parallel_actions(actionstr_input: Union[List[str], str], state: EnvStates) -> Tuple[bool, Optional[EnvStates]]:
    current_env = state.env 
    current_env.reset(states=state.cur_states) 
    
    action_to_simulate = actionstr_input
    if isinstance(actionstr_input, list):
        action_to_simulate = "\n".join(actionstr_input)

    out = current_env.simulate_one_step(action_to_simulate)
    
    if out["success"]:
        new_state_data = current_env.get_states()
        new_search_state = EnvStates(
            current_env,
            parent_state_data=state.cur_states, 
            current_state_data=new_state_data,
        )
        return True, new_search_state
    else:
        return False, None


def generate_potential_actions(state: EnvStates) -> List[Tuple[str, EnvStates]]:
    if not hasattr(state.env, 'robot_names'):
        return []
        
    robot_names = sorted(state.env.robot_names)

    single_robot_potential_actions: Dict[str, List[str]] = {
        r: generate_single_robot_action(r, state) for r in robot_names
    }

    valid_action_sets: List[Tuple[str, EnvStates]] = []
    action_verification_tasks = [] 

    for robot_id, actions in single_robot_potential_actions.items():
        for action in actions:
            action_verification_tasks.append(([action], state))

    active_robots = [r for r in robot_names if single_robot_potential_actions[r]]
    if len(active_robots) >= 2:
        max_concurrent_robots = 2
        if hasattr(state.env, 'object_names'): # Check if object_names exists before using its length
             max_concurrent_robots = min(4, len(state.env.object_names), len(active_robots))
        else: # Fallback if object_names is not available
             max_concurrent_robots = min(2, len(active_robots))


        for group_size in range(2, max_concurrent_robots + 1):
            if group_size > len(active_robots): continue
            for robot_group in itertools.combinations(active_robots, group_size):
                action_combos_for_group = list(
                    itertools.product(
                        *[single_robot_potential_actions[r] for r in robot_group]
                    )
                )
                cleaned_combos = [[a for a in combo] for combo in action_combos_for_group]
                for combo in cleaned_combos:
                    action_verification_tasks.append((combo, state))
    
    verify_results = []
    for action_combo, original_state_for_verification in action_verification_tasks:
        success, new_state_obj = verify_parallel_actions(action_combo, original_state_for_verification)
        if success and new_state_obj is not None:
            verify_results.append(("\n".join(action_combo), new_state_obj))

    verify_results.sort(
        key=lambda x_tuple: (x_tuple[1].heuristic(), -x_tuple[0].count("True"), -len(x_tuple[0].split("\n")))
    )
    
    return verify_results[:20]


def reconstruct_path(came_from: Dict[int, Tuple[Optional[int], str]], final_hash: int) -> List[str]:
    actions_sequence = []
    current_hash = final_hash
    while current_hash in came_from:
        parent_hash, action = came_from[current_hash]
        if action is not None: 
            actions_sequence.append(action)
        if parent_hash is None: 
            break
        current_hash = parent_hash
    actions_sequence.reverse()
    return actions_sequence

\end{fancyminted}

\section{\boxnettwodsp and \boxnetthreedsp Environment Examples}\label{sec:appendix-example}
In this section, we provide more examples of the two environments we developed in this work. We note that all examples shown in this section are in $3\times 3$ and $4 \times 4$ grid maps, but the dataset contains a wider range of map sizes. For more \boxnetthreedsp video examples, please visit our project website at this anonymous link \website. 

\subsection{\boxnettwodsp Examples}
\begin{figure}[h!]
    \centering
    \includegraphics[width=0.45\linewidth]{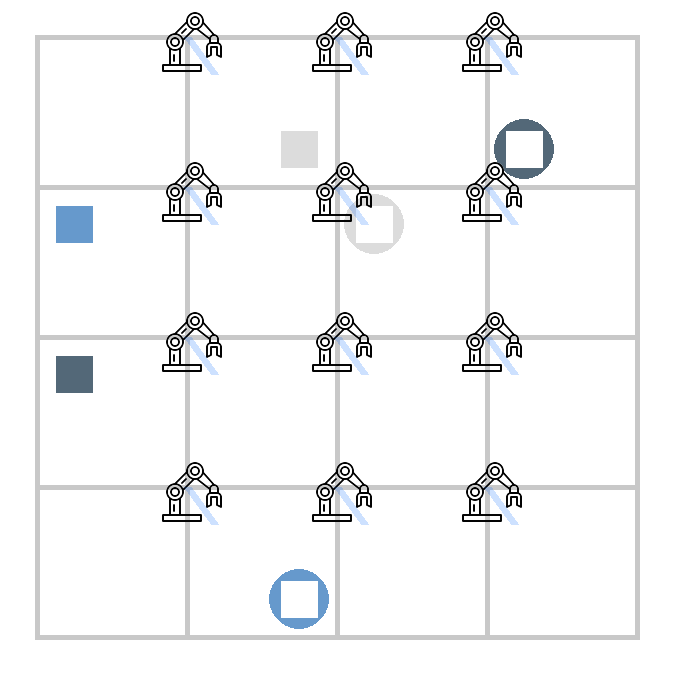}
    \includegraphics[width=0.45\linewidth]{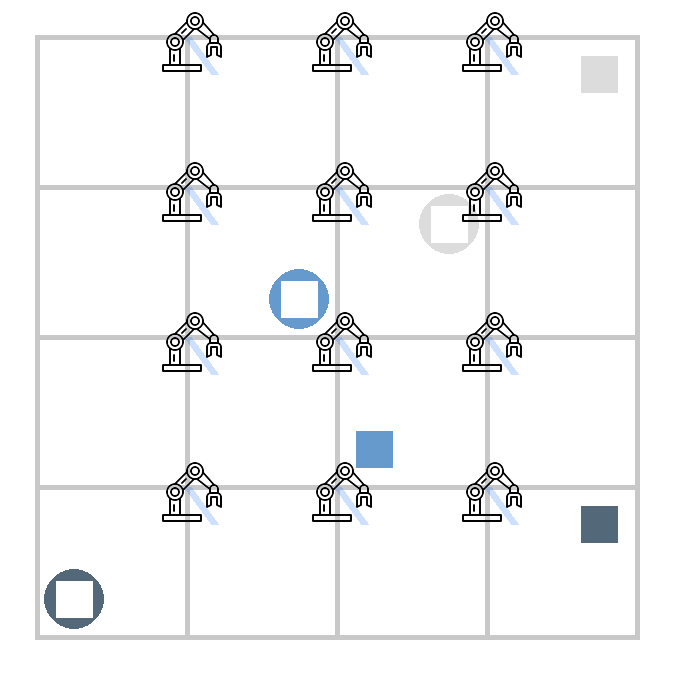}
    \caption{Example \boxnettwodsp environment}
    \label{fig:box2d-example}
\end{figure}

\begin{figure}[h!]
    \centering
    \includegraphics[width=0.45\linewidth]{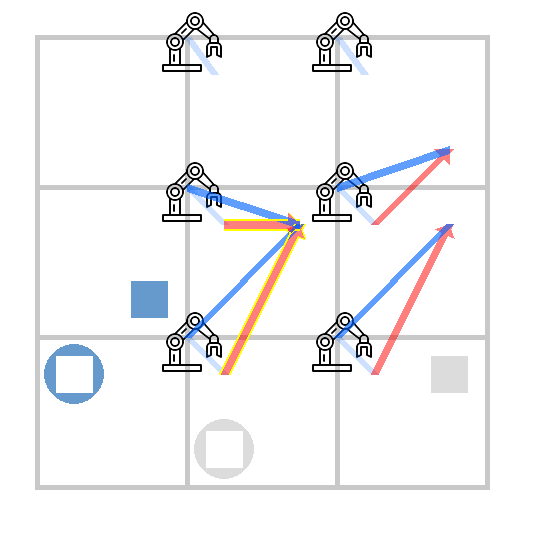}
    \includegraphics[width=0.45\linewidth]{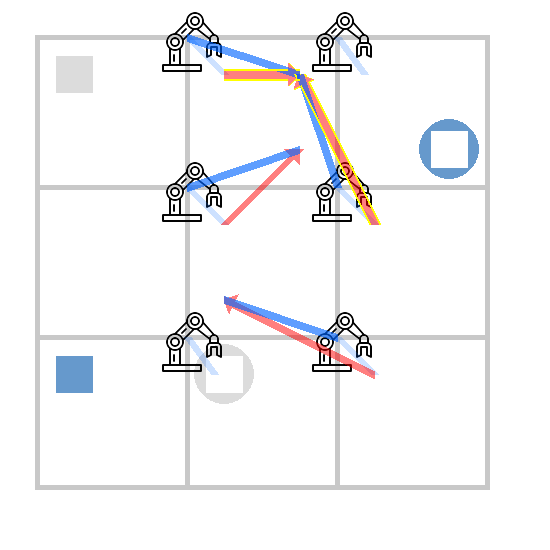} \\
    \includegraphics[width=0.45\linewidth]{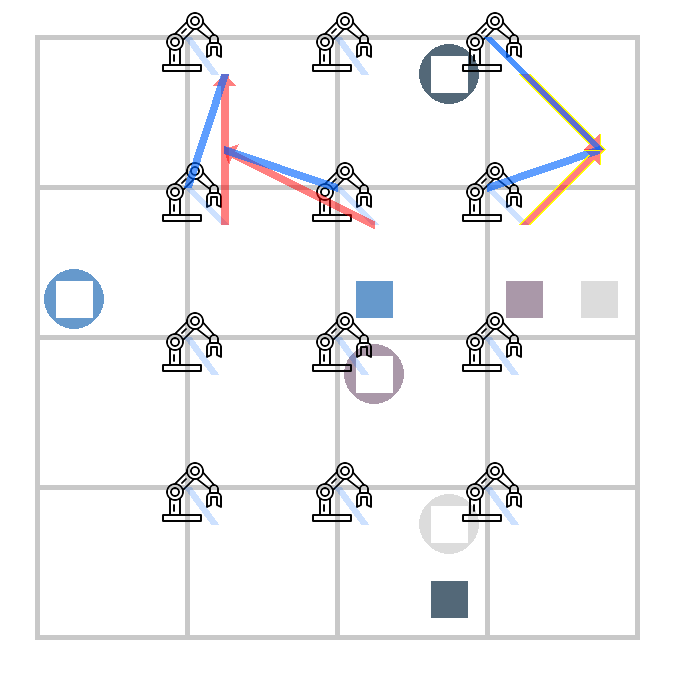}
    \includegraphics[width=0.45\linewidth]{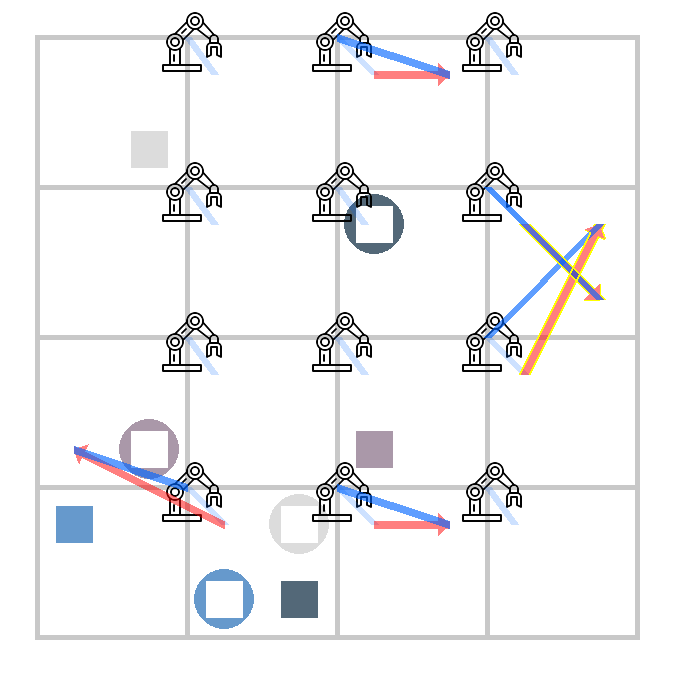}
    \caption{Example collisions in \boxnettwodsp environment. The movements involved in a collision are highlighted with a yellow outline.}
    \label{fig:box2d-collision}
\end{figure}

\paragraph{Original \boxnettwodsp examples}
Figure~\ref{fig:box2d-example} shows two example \boxnettwodsp environments.
Figure~\ref{fig:box2d-collision} shows four example collisions in \boxnettwodsp environments.

\begin{figure}[h!]
    \centering
    \includegraphics[width=0.45\linewidth]{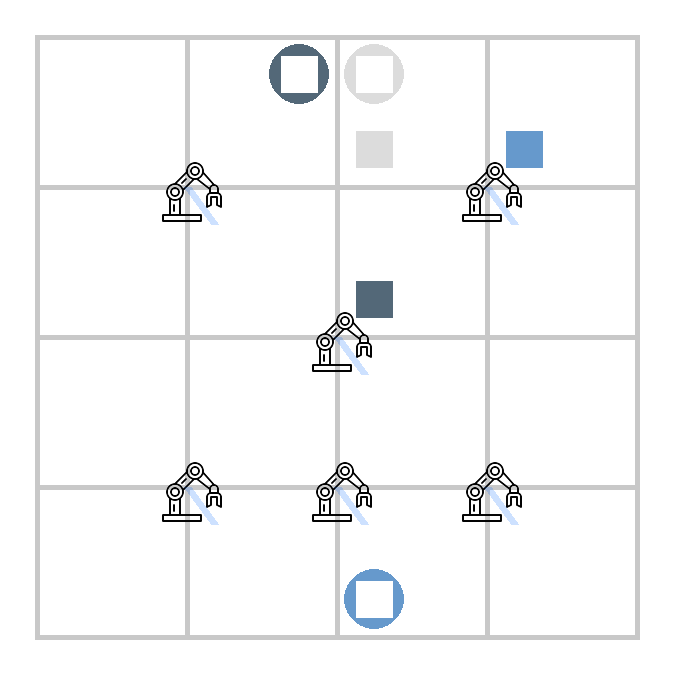}
    \includegraphics[width=0.45\linewidth]{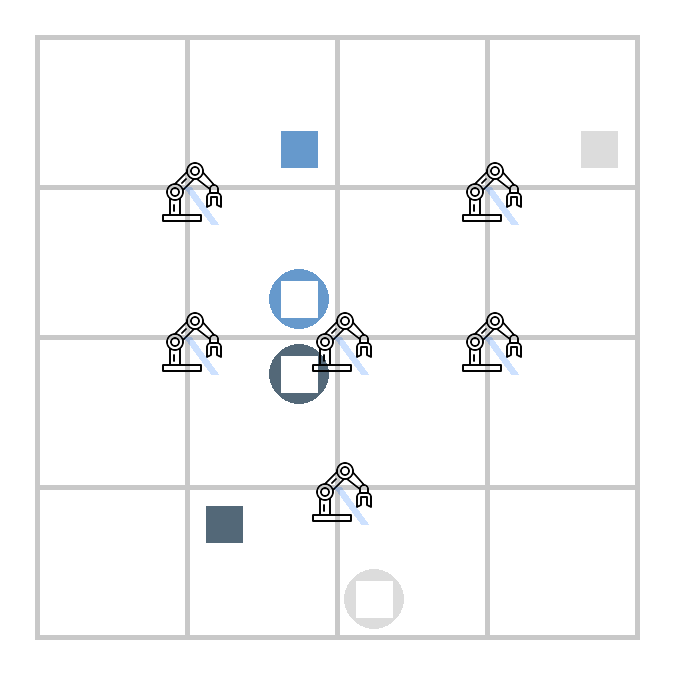}
    \caption{Example \textsc{RandRob} environment.}
    \label{fig:box2d-randrob}
\end{figure}

\begin{figure}[h!]
    \centering
    \includegraphics[width=0.45\linewidth]{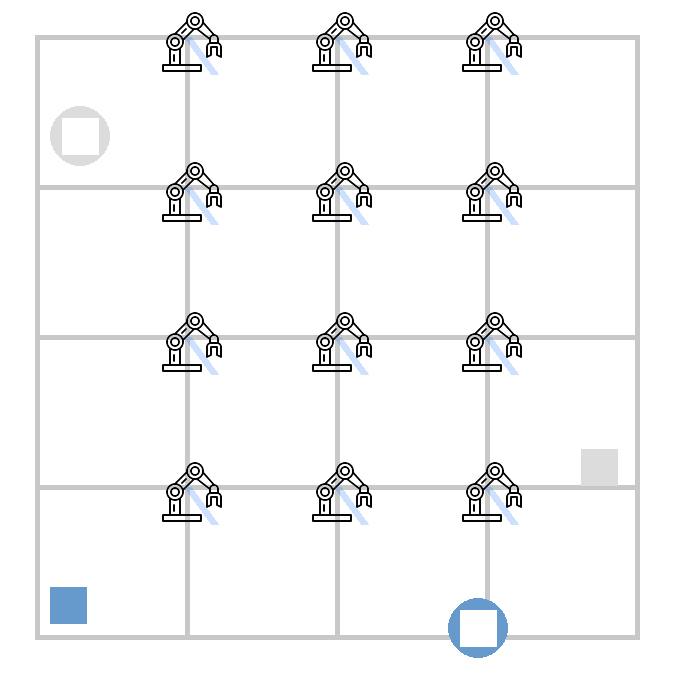}
    \includegraphics[width=0.45\linewidth]{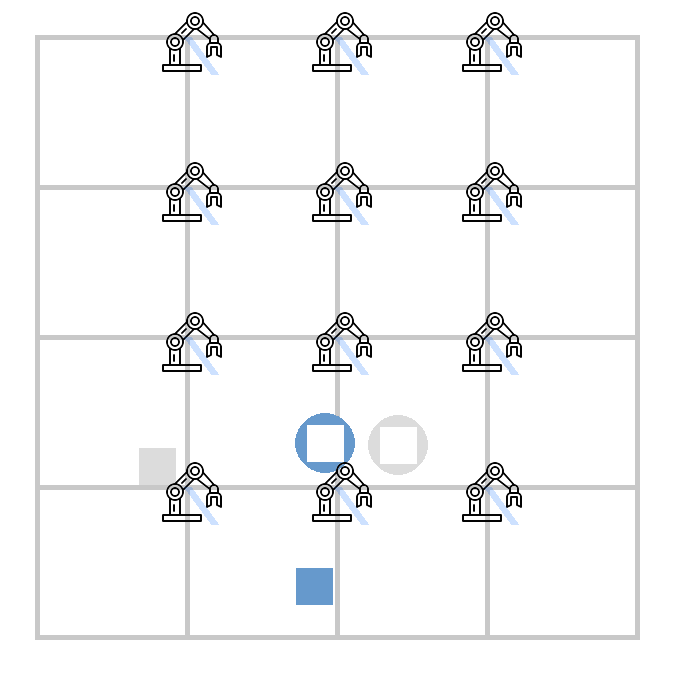}
    \caption{Example \textsc{NewCoord} environment.}
    \label{fig:box2d-newcoord}
\end{figure}

\paragraph{Unseen \boxnettwodsp examples for generalization experiment}
Figure~\ref{fig:box2d-randrob} shows two example \textsc{RandRob}  environments. Figure~\ref{fig:box2d-newcoord} shows two example \textsc{NewCoord} environments.


\subsection{\boxnetthreedsp Examples}
\begin{figure}[h!]
    \centering
    \includegraphics[width=0.9\linewidth]{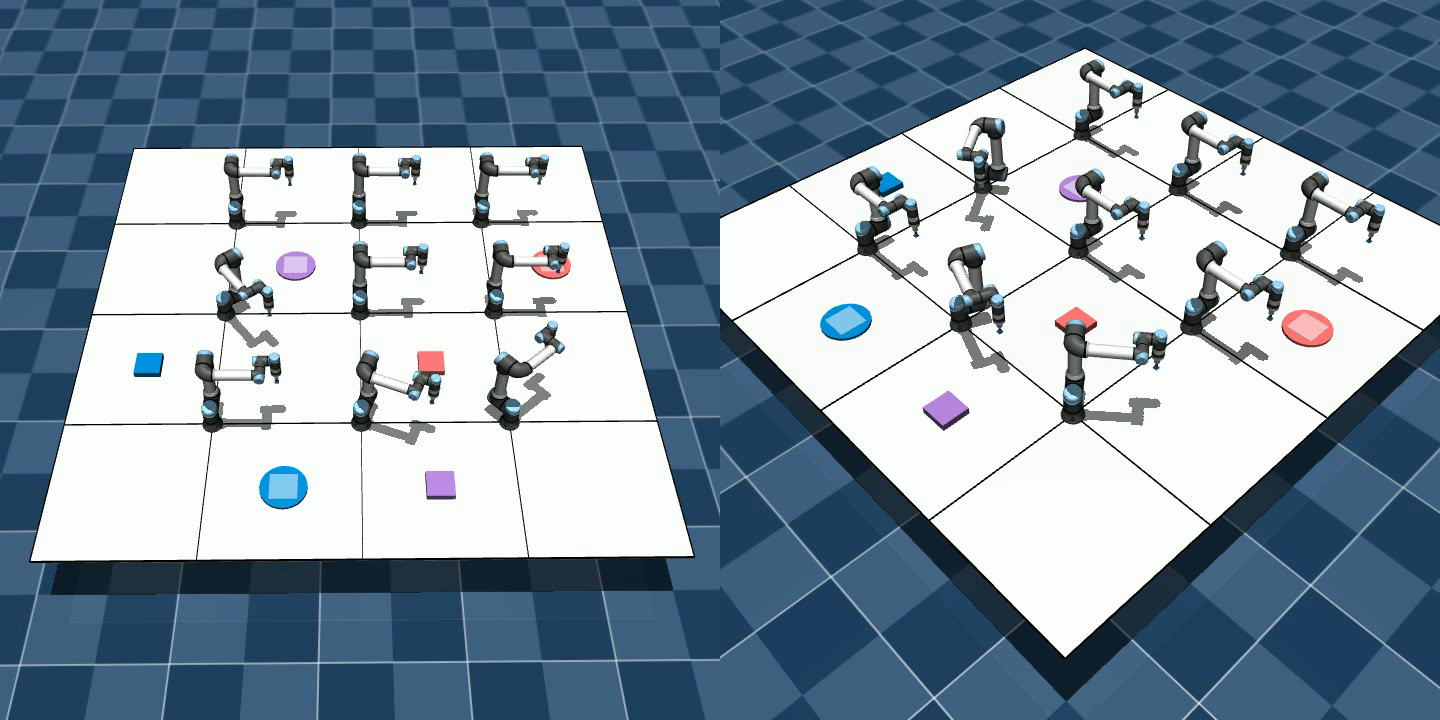} \\
    \includegraphics[width=0.9\linewidth]{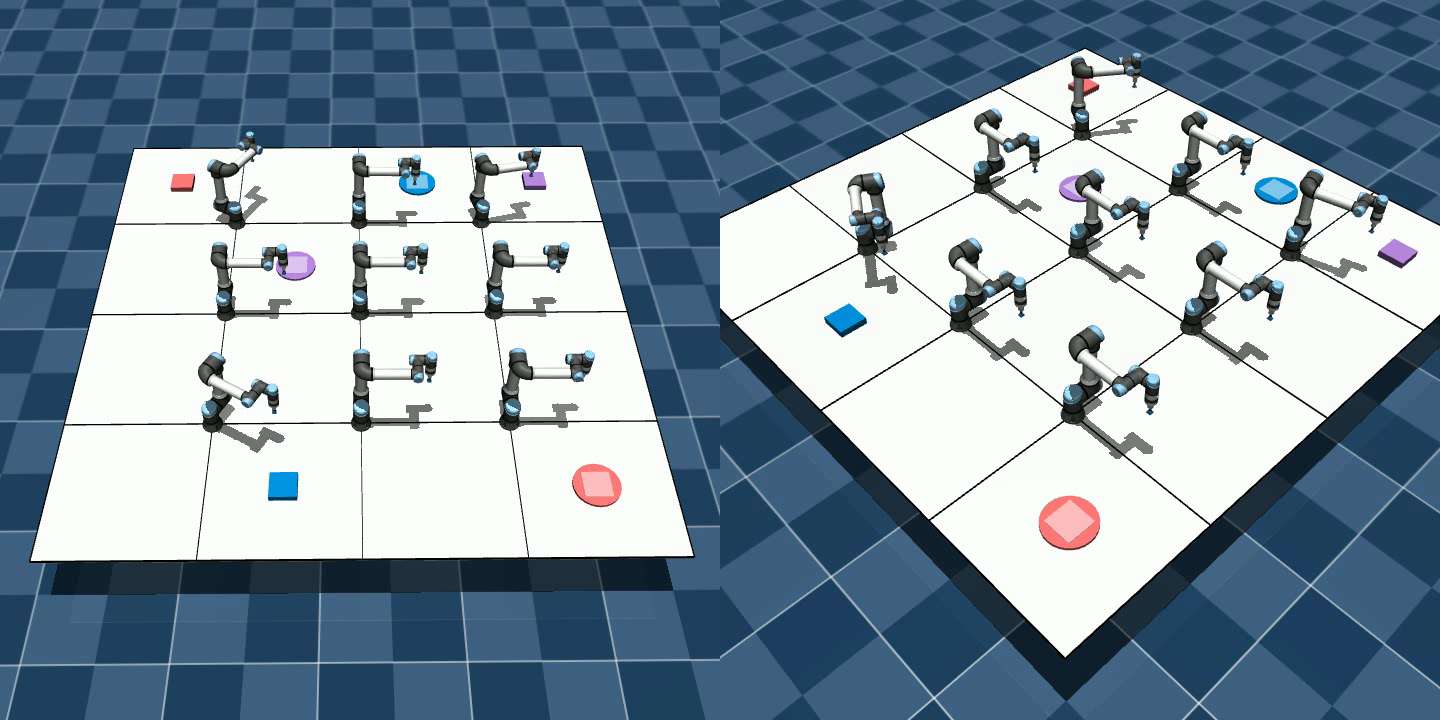}
    \caption{Example \boxnetthreedsp environment.}
    \label{fig:box3d-example}
\end{figure}

Figure~\ref{fig:box3d-example} shows examples for \boxnetthreedsp environments.

\clearpage
\section{Detailed Prompts}\label{sec:appendix-prompts}

In this section, we summarize the full prompts for SFT data synthesis in Section~\ref{subsec:sft-prompt}, \boxnettwodsp and \boxnetthreedsp environments with two planner modes in Section~\ref{subsec:environment-prompt}, and reasoning behavior analysis in Section~\ref{subsec:checknumber-prompt}.

\subsection{Prompt for SFT Data Synthesis}\label{subsec:sft-prompt}

We list the prompt for \boxnettwodsp thinking synthesis in Listing~\ref{lst:synth2dprompt}, and \boxnetthreedsp thinking synthesis in Listing~\ref{lst:synth3dprompt}.
\begin{fancyminted}{markdown}{Prompt for synthesizing reasoning trace for \boxnettwod}{lst:synth2dprompt}
You are required to **assume the role of a central planner**. Your task is to simulate the step-by-step thinking process that logically leads you to the provided ground-truth plan.

Your thinking should be presented from a **first-person perspective**, clearly demonstrating your internal reasoning process of planning, validating and adjusting to avoid collision, and planning decisions. 

## Requirement for your generated firt-person thinking:
1. **First-Person Perspective**: Write your internal thoughts as if you are personally making the decisions:
   - Use phrases like "Let me see...", "Wait, is that correct?", "I should check collisions first...", "Can I parallel two robot movements to make the plan more efficient?"
   - Demonstrate real-time analysis and potential hesitations or reconsiderations.
2. **Thinking Process with `<think>` Tags**:
   - Enclose your entire reasoning sequence in `<think>` ... `</think>` tags.
   - Make sure you have explicit checks, e.g. collision checks, range feasibility, and confirmations of correctness. You can start the explicit checks with "Wait", "Hmm", "let me check", etc.
   - Make sure to pose questions to yourself, and then answer them. Show how you arrive at each movement decision.
   - You must include multiple explict checks and self-questioning in your thinking process.


Below is the detailed task description. You can learn the rules for the task from these descriptions. 
   
## Task Description:
You are a central planner responsible for coordinating multiple robotic arms operating in a grid-like environment. Your goal is to plan and execute efficient, collision-free movements to transport objects to their designated target positions.


*Task Representation:*
* Objective: Move all objects to their specified target locations safely and efficiently.
* Input: A detailed map state containing positions of robots, objects, and target locations.
* Output: A precise movement plan specifying each robot arm's actions for moving objects.


*Position Representation:*
* All positions (robots, objects, targets) are given by their center coordinates, e.g., [0.25, 0.25], [0.75, 1.25].
* Robots have a fixed base location and an extendable arm with a limited reach range.


*Movement Rules:*
* Each robot arm can only move within a limited range relative to its fixed base position:
    * X-axis: from (Base_X - 1.0) to (Base_X + 1.0) (exclusive).
    * Y-axis: from (Base_Y - 1.0) to (Base_Y + 1.0) (exclusive).
* For example:
    * If a robot's base is [1.0, 1.0], its arm can reach [0.25, 0.75] or [1.25, 1.75], but not [0, 0.25] or [2.0, 0.75].
    * Robots may move an object only if their arm aligns exactly with the object's current position, and if explicitly indicated in the action (move_object: True).


## How to Generate Your Response:
Your response must **clearly indicate your thinking process** enclosed in <think> and </think> tags, followed by the generated step of the movement plan.

*Thinking:*
* Clearly describe your analysis and decisions from a first-person perspective.
* Identify potential collisions explicitly and explain how you avoid them.
* Highlight your reasoning for movement choices, considering efficiency and collision avoidance.

*Movement Plan (Output):*
* Your generated step of the movement plan should be in markdown format and contain a JSON dictionary, with robot names as keys and their movement instructions as values, structured as follows:
```json
[
{
    "robot_name": "start_position -> end_position, move_object"
    "robot_name": "start_position -> end_position, move_object",
},
{
    "robot_name": "start_position -> end_position, move_object",
}
]
```
* *start_position* and *end_position* represent the *[x, y]* coordinates of the robot arm's movement.
* *move_object* is a boolean indicating whether the robot moves an object (*True*) or simply moves its arm without carrying an object (*False*).
* Robots without actions in the current step should not be included.
Ensure your final step completes the objective of placing all objects at their target positions, and your plan forms a valid JSON array.


## Collision Avoidance Rules:
Your plan must strictly avoid collisions, as follows:
* Robot-Robot Collision:
    * Two robot arms cannot occupy the same position simultaneously.
    * Robot arms cannot cannot intersect with each other or have intersecting movement trajectories during a step movement.
    * For example:
        * Collision occurs if Robot 1 moves [0.75, 0.75] -> 0.75, 1.25] and Robot 2 moves [2.25, 1.75] -> [0.75, 1.25] (same endpoint).
        * Collision occurs if Robot 1 moves [0.25, 0.25] -> [0.75, 0.25] and Robot 2 moves [1.25, 0.25] -> [0.25, 0.75] (intersecting arms as the end position Robot 1 is at the arm, as the end of Robot 2 arm position occupies [0.75, 0.25])
        * Collision occurs if Robot 1 moves [0.25, 0.25] -> [0.75, 0.75] and Robot 2 moves [0.25, 0.75] -> [0.75, 0.25] (intersecting movement as both arms moves across [0.5, 0.5]).
    * Object-Object Collision:
        * Two objects cannot occupy the same position at any time.

## Example Environment and Ground-Truth Plan:
Below is an example scenario and its ground-truth solution:

```text
{environment}
```

With the above information clearly provided, please start by explicitly presenting your first-person reasoning for the whole plan enclosed in <think></think> tags. Make sure you include explicit checks and self-questioning in your thinking process. Your reasoning should be clear and easy to follow, as if you are explaining it to someone else. Limit your thinking length within 2000 tokens.
\end{fancyminted}

\begin{fancyminted}{markdown}{Prompt for synthesizing reasoning trace for \boxnetthreed}{lst:synth3dprompt}
You are required to **assume the role of a central planner**. Your task is to simulate the step-by-step thinking process that logically leads you to the provided ground-truth movement plan.

Your thinking should be presented from a **first-person perspective**, clearly demonstrating your internal reasoning process of planning, validating and adjusting to avoid collision, and planning decisions. 

## Requirement for your generated firt-person thinking:
1. **First-Person Perspective**: Write your internal thoughts as if you are personally making the decisions:
   - Use phrases like "Let me see...", "Wait, is that correct?", "I should check collisions first...", "Can I parallel two robot movements to make the plan more efficient?"
   - Demonstrate real-time analysis and potential hesitations or reconsiderations.
2. **Thinking Process with `<think>` Tags**:
   - Enclose your entire reasoning sequence in `<think>` ... `</think>` tags.
   - Make sure you have explicit checks, e.g. collision checks, range feasibility, and confirmations of correctness. You can start the explicit checks with "Wait", "Hmm", "let me check", etc.
   - Make sure to pose questions to yourself, and then answer them. Show how you arrive at each movement decision.
   - You must include multiple explict checks and self-questioning in your thinking process.


Below is the detailed task description. You can learn the rules for the task from these descriptions. 
   
## Task Description:
You are a central planner responsible for coordinating multiple robotic arms operating in a grid-like environment. Your goal is to plan and execute efficient, collision-free movements to transport objects to their designated target positions.


*Task Representation:*
* Objective: Move all objects to their specified target locations safely and efficiently.
* Input: A detailed map state containing positions of robots, objects, and target locations.
* Output: A precise movement plan specifying each robot arm's actions for moving objects.


*Position Representation:*
* All positions (robots, objects, targets) are given by their center coordinates, e.g., [0.55, 1.65], [2.75, 0.55].
* Robots have a fixed base location and an extendable arm with a limited reach range.


*Movement Rules:*
* Each robot arm can only move within a circular band around its fixed base position:  
  - Let d = sqrt((X - Base_X)**2 + (Y - Base_Y)**2).  
  - The arm may reach (X, Y) only if 0.4 < d < 0.8

* For example:  
  - If a robot's base is at [1.1, 1.1]:  
    - It can reach [0.55, 0.55] since sqrt((1.1 - 0.55)**2 + (1.1 - 0.55)**2) around 0.77 < 0.8
    - It can reach [0.6, 1.1] since sqrt((0.6 - 1.1)**2 + (1.1 - 1.1) ** 2)) = 0.5 > 0.4
    - It cannot reach [2.0, 1.1] because sqrt(0.9**2 + 0**2) = 0.9, which exceeds 0.8
    - It cannot reach [2.25, 0.65] because sqrt(1.15**2 + 0.45**2) around 1.23, which exceeds 0.8
  - If a robot needs to move an object within its range and the arm is not aligned with the object, the robot should first move its arm to the position of that object. By aligning, it meas the distance between object center and arm position is less than 0.1
  - When you plan a move, please follow following rules:
    - First check that the proposed target lies within the circular band 0.5 < d < 0.8.  
    - If it does not, adjust your plan or reject that movement.  
    - If the arm is not yet aligned with an object it needs to move and that object lies within the band, plan a preliminary move to position the arm aligned with the object before picking it up.



## How to Generate Your Response:
Your response must **clearly indicate your thinking process** enclosed in <think> and </think> tags, followed by the generated step of the movement plan.

*Thinking:*
* Clearly describe your analysis and decisions from a first-person perspective.
* Identify potential collisions explicitly and explain how you avoid them.
* Highlight your reasoning for movement choices, considering efficiency and collision avoidance.

*Movement Plan (Output):*
* Your generated step of the movement plan should be in markdown format and contain a JSON dictionary, with robot names as keys and their movement instructions as values, structured as follows:
```json
[
{
    "robot_name1": "Move end_position, move_object",
    "robot_name2": "Move end_position, move_object"
},
{
    "robot_name3": "Move end_position, move_object"
}
]
```
* *end_position* represent the target *[x, y]* coordinates of the robot arm end point of the movement around circular path. Note that only the arm moves while its base remains fixed.
* *move_object* is a boolean indicating whether the robot moves an object (*True*) or simply moves its arm without carrying an object (*False*).
* One robot can only be moved once in each step, which means that no repeated keys are allowed in the same step.
* Robots without actions in the current step should not be included.
Ensure your final step completes the objective of placing all objects at their target positions, and your plan forms a valid JSON array.


## Collision Avoidance Rules:
Your plan must strictly avoid collisions, as follows:
* Robot-Robot Collision
  * Each robot arm always swings along a smooth **circular** path around its base.
  * Two robot arms cannot occupy the same position at the end of a move.  
  * Their curved paths must not cross or share any point during the move.  
  * Sometimes a robot needs to move its arm to a safe position to avoid collision between another robot that move its arm to reach an object.
  * **Example:**  
    * robot_0 swings from [0.25, 0.25] to [0.75, 0.75] and robot_1 swings from [0.25, 0.75] to [0.75, 0.25] at the same time. Both arcs pass through [0.5, 0.5], causing a collision.
* Object-Object Collision
  * Two objects cannot occupy the same (x, y) at any time.  
  * If you move more than one object at once, they must have different drop-off points and non-crossing straight-line paths.
* Robot-Object Collision
  * An arm's circular path must not sweep through any object it isn't carrying.  
  * Before moving, confirm the curved trajectory does not pass over another object's position.


## Plan Efficiency Considerations:
* Each step of your plan involves simultaneous robot arm movements from their current positions to specified target positions.
* Each robot arm moves at a constant speed of 0.5 units/time.
* The duration of each step is determined by the longest single-arm movement within that step.
* The total execution time is the sum of all individual step durations.
* You should aim to minimize total execution time while ensuring collision-free movements and successful object placements.

## Example Environment and Ground-Truth Plan:
Below is an example scenario and its ground-truth solution:

```text
{environment}
```

With the above information clearly provided, please start by explicitly presenting your first-person reasoning for the whole plan enclosed in <think></think> tags. Make sure you include explicit checks and self-questioning in your thinking process. Your reasoning should be clear and easy to follow, as if you are explaining it to someone else. Limit your thinking length within 2000 tokens.
\end{fancyminted}

\subsection{Prompt for \boxnettwodsp and \boxnetthreedsp environment}\label{subsec:environment-prompt}

\paragraph{\boxnettwod} 
We list the prompt for \fullplansp planner in \boxnettwodsp in List~\ref{lst:fullplanprompt2d}, and \replansp planner in List~\ref{lst:replanprompt2d}.

\paragraph{\boxnetthreed} 
We list the prompt for \fullplansp planner in \boxnetthreedsp in List~\ref{lst:fullplanprompt3d}, and \replansp planner in List~\ref{lst:replanprompt3d}.

\begin{fancyminted}{markdown}{Prompt for \fullplansp planner in \boxnettwodsp environment}{lst:fullplanprompt2d}

You are a central planner responsible for coordinating robotic arms in a grid-like environment to transport objects to their designated targets. Each robot is stationed at the corner of a 1x1 square and uses its arm to move objects. Your task is to generate an efficient and collision-free plan for multiple robots, ensuring all objects reach their target positions after the whole plan is executed.


## Task Description:
*Task Representation:*
* Objective: Move all objects to their specified target locations safely and efficiently.
* Input: A detailed map state containing positions of robots, objects, and target locations.
* Output: A precise movement plan specifying each robot arm's actions for moving objects.


*Position Representation:*
* All positions (robots, objects, targets) are given by their center coordinates, e.g., [0.25, 0.25], [0.75, 1.25].
* Robots have a fixed base location and an extendable arm with a limited reach range.


*Movement Rules:*
Your generated movement must strictly consider the reachability of each robot arm, detaild rule in following:
* Each robot arm can only move within a limited range relative to its fixed base position:
    * X-axis: from (Base_X - 1.0) to (Base_X + 1.0) (exclusive).
    * Y-axis: from (Base_Y - 1.0) to (Base_Y + 1.0) (exclusive).
* For example:
    * If a robot's base is [1.0, 1.0], its arm can reach [0.25, 0.75] or [1.25, 1.75], but not [0, 0.25] or [2.25, 1.75] because 2.25 - 1.0 = 1.25 > 1.0 and 0 - 1.0 = -1.0 <= -1.
    * Robots may move an object only if their arm position aligns exactly with the object's current position, and if explicitly indicated in the action (move_object: True).
    * Make sure you explicitly think about whether your proposed movement for one arm is valid, and correct it if it is not.
    * If a robot needs to move an object within its range and the arm is not aligned with the object, the robot should first move its arm to the position of that object.


## How to Generate Your Response:
Your response must **clearly indicate your thinking process** enclosed in <think> and </think> tags, followed by the generated step of the movement plan.


*Thinking:*
* Clearly describe your analysis and decisions from a first-person perspective.
* You should think carefully whether your plan has collision by explictly generating your thoughts, and avoid them in your final output if there is any.
* Highlight your reasoning for movement choices, considering efficiency and collision avoidance.

*Movement Plan (Output):*
* Your generated step of the movement plan should be in markdown format and contain a JSON list, with each entry as a dictionary indicating one step, and the robot names are keys and their movement instructions as values for each step, structured as follows:
```json
[
{
    "robot_name1": "start_position -> end_position, move_object",
    "robot_name2": "start_position -> end_position, move_object"
},
{
    "robot_name3": "start_position -> end_position, move_object"
}
]
```
* *start_position* and *end_position* represent the *[x, y]* coordinates of the robot arm before and after the movement. Note that only the arm moves while base remains fixed.
* *move_object* is a boolean indicating whether the robot moves an object (*True*) or simply moves its arm without carrying an object (*False*).
* Robots without actions in a certain step should not be included.
* One robot can only be moved once in each step, which means that no repeated keys are allowed in the same step.
Ensure your final step completes the objective of placing all objects at their target positions, and your plan forms a valid JSON list.


## Collision Avoidance Rules:
Your plan must strictly avoid collisions, as follows:
* Robot-Robot Collision:
    * Two robot arms cannot occupy the same position simultaneously.
    * Robot arms cannot cannot intersect with each other or have intersecting movement trajectories during a step movement.
    * For example:
        * Collision occurs if Robot 1 moves [0.75, 0.75] -> 0.75, 1.25] and Robot 2 moves [2.25, 1.75] -> [0.75, 1.25] (same endpoint).
        * Collision occurs if Robot 1 moves [0.25, 0.25] -> [0.75, 0.25] and Robot 2 moves [1.25, 0.25] -> [0.25, 0.75] (intersecting arms as the end position Robot 1 is at the arm, as the end of Robot 2 arm position occupies [0.75, 0.25])
        * Collision occurs if Robot 1 moves [0.25, 0.25] -> [0.75, 0.75] and Robot 2 moves [0.25, 0.75] -> [0.75, 0.25] (intersecting movement as both arms moves across [0.5, 0.5]).
    * Object-Object Collision:
        * Two objects cannot occupy the same position at any time.


## Plan Efficiency Considerations:
* Each step of your plan involves simultaneous robot arm movements from their current positions to specified target positions.
* Each robot arm moves at a constant speed of 0.5 units/time.
* The duration of each step is determined by the longest single-arm movement within that step.
* The total execution time is the sum of all individual step durations.
* You should aim to minimize total execution time while ensuring collision-free movements and successful object placements.


## Example Input & Output:
* Input:
Object positions:
    Object 1: [0.75, 0.75]
    Object 2: [1.75, 0.25]
Target positions:
    Object 1 target: [2.25, 0.75]
    Object 2 target: [0.25, 1.25]
Robot positions:
    Robot 1: base [1.0, 1.0], arm [0.75, 0.75]
    Robot 2: base [2.0, 0.0], arm [1.75, 0.75]

* Output:
<think> Let's understand the scenerio ... </think>
```json
[
    {"Robot 1": "[0.75, 0.75] -> [1.25, 0.25], True", "Robot 2": "[1.75, 0.75] -> [1.75, 0.25], False"},
    {"Robot 2": "[1.75, 0.25] -> [1.75, 0.75], True"},
    {"Robot 1": "[1.25, 0.25] -> [1.75, 0.75], False", "Robot 2": "[1.75, 0.75] -> [1.25, 0.25], False"},
    {"Robot 1": "[1.75, 0.75] -> [0.25, 1.25], True", "Robot 2": "[1.25, 0.25] -> [2.25, 0.75], True"}
]
```

Given the information above, now consider the following environment:
Input:

{mapstate}

Generate the full plan for moving these robots. 
\end{fancyminted}

\begin{fancyminted}{markdown}{Prompt for \replansp planner in \boxnettwodsp environment}{lst:replanprompt2d}
You are a central planner responsible for coordinating robotic arms in a grid-like environment to transport objects to their designated targets. Each robot is stationed at the corner of a 1x1 square and uses its arm to move objects. Your task is to interactively generate an efficient and collision-free movement plan for controlling these robots, targeting at moving all objects to their target positions. 

At each step, you will receive the current state of the environment wrapped by <observation> and </observation> tags. You need to generate the next-step plan for moving the robots, ensuring that your output contains your thinking process and the markdown json dict.


## Task Description:
*Task Representation:*
* Objective: Move all objects to their specified target locations safely and efficiently.
* Input: A detailed map state containing positions of robots, objects, and target locations.
* Output: A precise movement plan specifying each robot arm's actions for moving objects.


*Position Representation:*
* All positions (robots, objects, targets) are given by their center coordinates, e.g., [0.25, 0.25], [0.75, 1.25].
* Robots have a fixed base location and an extendable arm with a limited reach range.


*Movement Rules:*
* Each robot arm can only move within a limited range relative to its fixed base position:
    * X-axis: from (Base_X - 1.0) to (Base_X + 1.0) (exclusive).
    * Y-axis: from (Base_Y - 1.0) to (Base_Y + 1.0) (exclusive).
* For example:
    * If a robot's base is [1.0, 1.0], its arm can reach [0.25, 0.75] or [1.25, 1.75], but not [0, 0.25] or [2.25, 1.75] because 2.25 - 1.0 = 1.25 > 1.0 and 0 - 1.0 = -1.0 <= -1.
    * Robots may move an object only if their arm aligns exactly with the object's current position, and if explicitly indicated in the action (move_object: True).
    * If a robot needs to move an object within its range and the arm is not aligned with the object, the robot should first move its arm to the position of that object.


## How to Generate Your Response:
Your response must **clearly indicate your thinking process** enclosed in <think> and </think> tags, followed by the generated step of the movement plan.


*Thinking:*
* Clearly describe your analysis and decisions from a first-person perspective.
* Think carefully and try to identify potential collisions explicitly in the analysis and explain how you avoid them.
* Highlight your reasoning for movement choices, considering efficiency and collision avoidance.


*Movement Plan (Output):*
* Your generated step of the movement plan should be in markdown format and contain a JSON dictionary, with robot names as keys and their movement instructions as values, structured as follows:
```json
{
    "robot_name1": "start_position -> end_position, move_object"
    "robot_name2": "start_position -> end_position, move_object",
},
```
* *start_position* and *end_position* represent the *[x, y]* coordinates of the robot arm before and after the movement. Note that only the arm moves while base remains fixed.
* *move_object* is a boolean indicating whether the robot moves an object (*True*) or simply moves its arm without carrying an object (*False*).
* Robots without actions in a certain step should not be included.
* One robot can only be moved once in each step, which means that no repeated keys are allowed in the same step.
Ensure your output forms a valid JSON dictionary of next-step plan.



## Collision Avoidance Rules:
Your plan must strictly avoid collisions, as follows:
* Robot-Robot Collision:
    * Two robot arms cannot occupy the same position simultaneously.
    * Robot arms cannot cannot intersect with each other or have intersecting movement trajectories during a step movement.
    * For example:
        * Collision occurs if Robot 1 moves [0.75, 0.75] -> 0.75, 1.25] and Robot 2 moves [2.25, 1.75] -> [0.75, 1.25] (same endpoint).
        * Collision occurs if Robot 1 moves [0.25, 0.25] -> [0.75, 0.25] and Robot 2 moves [1.25, 0.25] -> [0.25, 0.75] (intersecting arms as the end position Robot 1 is at the arm, as the end of Robot 2 arm position occupies [0.75, 0.25])
        * Collision occurs if Robot 1 moves [0.25, 0.25] -> [0.75, 0.75] and Robot 2 moves [0.25, 0.75] -> [0.75, 0.25] (intersecting movement as both arms moves across [0.5, 0.5]).
* Object-Object Collision:
    * Two objects cannot occupy the same position at any time.


## Plan Efficiency Considerations:
* The exeuction time of your plan involves simultaneous robot arm movements from their current positions to specified target positions.
* Each robot arm moves at a constant speed of 0.5 units/time.
* The duration of the plan is determined by the longest single-arm movement within it.
* You should aim to minimize the execution time while ensuring collision-free movements and successful object placements.


## Example Input & Output:
Input:
<observation>
Object positions:
    Object 1: [0.75, 0.75]
    Object 2: [1.75, 0.25]
Target positions:
    Object 1 target: [2.25, 0.75]
    Object 2 target: [0.25, 1.25]
Robot positions:
    Robot 1: base [1.0, 1.0], arm [0.75, 0.75]
    Robot 2: base [2.0, 0.0], arm [1.75, 0.75]
</observation>

* Output:
<think> Let's understand the scenerio ... </think>
```json
{"Robot 1": "[0.75, 0.75] -> [1.25, 0.25], True", "Robot 2": "[1.75, 0.75] -> [1.75, 0.25], False"}
```

Now work on the following problem given by user.

<observation>
{mapstate}
</observation>
\end{fancyminted}
\begin{fancyminted}{markdown}{Prompt for \fullplansp planner in \boxnetthreedsp environment}{lst:fullplanprompt3d}
You are a central planner responsible for coordinating robotic arms in a grid-like environment to transport objects to their designated targets. Each robot is stationed at the corner of a 1.1x1.1 square and uses its arm to move objects. Your task is to generate an efficient and collision-free plan for multiple robots, ensuring all objects reach their target positions after the whole plan is executed.


## Task Description:
*Task Representation:*
* Objective: Move all objects to their specified target locations safely and efficiently.
* Input: A detailed map state containing positions of robots, objects, and target locations.
* Output: A precise movement plan specifying each robot arm's actions for moving objects.


*Position Representation:*
* All positions (robots, objects, targets) are given by their center coordinates, e.g., [0.55, 1.65], [2.75, 0.55].
* Robots have a fixed base location and an extendable arm with a limited reach range.


*Movement Rules:*
Your generated movement must strictly consider the reachability of each robot arm, detaild rule in following:
* Each robot arm can only move within a circular band around its fixed base position:  
  - Let d = sqrt((X - Base_X)**2 + (Y - Base_Y)**2).  
  - The arm may reach (X, Y) only if 0.4 < d < 0.8

* For example:  
  - If a robot's base is at [1.1, 1.1]:  
    - It can reach [0.55, 0.55] since sqrt((1.1 - 0.55)**2 + (1.1 - 0.55)**2) around 0.77 < 0.8
    - It can reach [0.6, 1.1] since sqrt((0.6 - 1.1)**2 + (1.1 - 1.1) ** 2)) = 0.5 > 0.4
    - It cannot reach [2.0, 1.1] because sqrt(0.9**2 + 0**2) = 0.9, which exceeds 0.8
    - It cannot reach [2.25, 0.65] because sqrt(1.15**2 + 0.45**2) around 1.23, which exceeds 0.8
  - If a robot needs to move an object within its range and the arm is not aligned with the object, the robot should first move its arm to the position of that object. By aligning, it meas the distance between object center and arm position is less than 0.1
  - When you plan a move, please follow following rules:
    - First check that the proposed target lies within the circular band 0.5 < d < 0.8.  
    - If it does not, adjust your plan or reject that movement.  
    - If the arm is not yet aligned with an object it needs to move and that object lies within the band, plan a preliminary move to position the arm aligned with the object before picking it up.


## How to Generate Your Response:
Your response must **clearly indicate your thinking process** enclosed in <think> and </think> tags, followed by the generated step of the movement plan.


*Thinking:*
* Clearly describe your analysis and decisions from a first-person perspective.
* You should think carefully whether your plan has collision by explictly generating your thoughts, and avoid them in your final output if there is any.
* Highlight your reasoning for movement choices, considering efficiency and collision avoidance.

*Movement Plan (Output):*
* Your generated step of the movement plan should be in markdown format and contain a JSON list, with each entry as a dictionary indicating one step, and the robot names are keys and their movement instructions as values for each step, structured as follows:
```json
[
{
    "robot_name1": "Move end_position, move_object",
    "robot_name2": "Move end_position, move_object"
},
{
    "robot_name3": "Move end_position, move_object"
}
]
```
* *end_position* represent the target *[x, y]* coordinates of the robot arm after the movement. Note that only the arm moves while its base remains fixed.
* *move_object* is a boolean indicating whether the robot moves an object (*True*) or simply moves its arm without carrying an object (*False*).
* Robots without actions in a certain step should not be included.
* One robot can only be moved once in each step, which means that no repeated keys are allowed in the same step.
Ensure your final step completes the objective of placing all objects at their target positions, and your plan forms a valid JSON list.


## Collision Avoidance Rules:
Your plan must strictly avoid collisions, as follows:
* Robot-Robot Collision
  * Each robot arm always swings along a smooth **circular** path around its base.
  * Two robot arms cannot occupy the same position at the end of a move.  
  * Their curved paths must not cross or share any point during the move.  
  * Sometimes a robot needs to move its arm to a safe position to avoid collision between another robot that move its arm to reach an object.
  * **Example:**  
    * robot_0 swings from [0.25, 0.25] to [0.75, 0.75] and robot_1 swings from [0.25, 0.75] to [0.75, 0.25] at the same time. Both arcs pass through [0.5, 0.5], causing a collision.
* Object-Object Collision
  * Two objects cannot occupy the same (x, y) at any time.  
  * If you move more than one object at once, they must have different drop-off points and non-crossing straight-line paths.
* Robot-Object Collision
  * An arm's circular path must not sweep through any object it isn't carrying.  
  * Before moving, confirm the curved trajectory does not pass over another object's position.


## Plan Efficiency Considerations:
* Each step of your plan involves simultaneous robot arm movements from their current positions to specified target positions.
* Each robot arm moves at a constant speed of 0.5 units/time.
* The duration of each step is determined by the longest single-arm movement within that step.
* The total execution time is the sum of all individual step durations.
* You should aim to minimize total execution time while ensuring collision-free movements and successful object placements.


## Example Input & Output:
* Input:
Object positions:
    Object 1: [0.55, 1.65]
Target positions:
    Object 1 target: [1.65, 0.55]
Robot positions:
    Robot 1: base [1.1, 1.1], arm [1.24, 0.61]
    Robot 2: base [1.1, 2.2], arm [1.24, 1.71]

* Output:
<think> Let's understand the scenerio ... </think>
```json
[
  {
    "Robot 1": "Move [0.55, 1.65] False"
  },
  {
    "Robot 1": "Move [1.65, 1.65] True"
  },
  {
    "Robot 1": "Move [1.10, 1.70] False", "Robot 0": "Move [1.65, 1.66] False"
  },
  {
    "Robot 0": "Move [1.65, 0.55] True"
  }
]
```

Given the information above, now consider the following environment:
Input:
{mapstate}
Generate the full plan for moving these robots.
\end{fancyminted}

\begin{fancyminted}{markdown}{Prompt for \replansp planner in \boxnetthreedsp environment}{lst:replanprompt3d}
You are a central planner responsible for coordinating robotic arms in a grid-like environment to transport objects to their designated targets. Each robot is stationed at the corner of a 1.1x1.1 square and uses its arm to move objects. Your task is to interactively generate an efficient and collision-free movement plan for controlling these robots, targeting at moving all objects to their target positions. 

At each step, you will receive the current state of the environment wrapped by <observation> and </observation> tags. You need to generate the next-step plan for moving the robots, ensuring that your output contains your thinking process and the markdown json dict.


## Task Description:
*Task Representation:*
* Objective: Move all objects to their specified target locations safely and efficiently.
* Input: A detailed map state containing positions of robots, objects, and target locations.
* Output: A precise movement plan specifying each robot arm's actions for moving objects.


*Position Representation:*
* All positions (robots, objects, targets) are given by their center coordinates, e.g., [0.55, 1.65], [2.75, 0.55].
* Robots have a fixed base location and an extendable arm with a limited reach range.


*Movement Rules:*
Your generated movement must strictly consider the reachability of each robot arm, detaild rule in following:
* Each robot arm can only move within a circular band around its fixed base position:  
  - Let d = sqrt((X - Base_X)**2 + (Y - Base_Y)**2).  
  - The arm may reach (X, Y) only if 0.4 < d < 0.8

* For example:  
  - If a robot's base is at [1.1, 1.1]:  
    - It can reach [0.55, 0.55] since sqrt((1.1 - 0.55)**2 + (1.1 - 0.55)**2) around 0.77 < 0.8
    - It can reach [0.6, 1.1] since sqrt((0.6 - 1.1)**2 + (1.1 - 1.1) ** 2)) = 0.5 > 0.4
    - It cannot reach [2.0, 1.1] because sqrt(0.9**2 + 0**2) = 0.9, which exceeds 0.8
    - It cannot reach [2.25, 0.65] because sqrt(1.15**2 + 0.45**2) around 1.23, which exceeds 0.8
  - If a robot needs to move an object within its range and the arm is not aligned with the object, the robot should first move its arm to the position of that object. By aligning, it meas the distance between object center and arm position is less than 0.1
  - When you plan a move, please follow following rules:
    - First check that the proposed target lies within the circular band 0.5 < d < 0.8.  
    - If it does not, adjust your plan or reject that movement.  
    - If the arm is not yet aligned with an object it needs to move and that object lies within the band, plan a preliminary move to position the arm aligned with the object before picking it up.



## How to Generate Your Response:
Your response must **clearly indicate your thinking process** enclosed in <think> and </think> tags, followed by the generated step of the movement plan.


*Thinking:*
* Clearly describe your analysis and decisions from a first-person perspective.
* Think carefully and try to identify potential collisions explicitly in the analysis and explain how you avoid them.
* Highlight your reasoning for movement choices, considering efficiency and collision avoidance.


*Movement Plan (Output):*
* Your generated step of the movement plan should be in markdown format and contain a JSON dictionary, with robot names as keys and their movement instructions as values, structured as follows:
```json
{
    "robot_name1": "Move end_position, move_object",
    "robot_name2": "Move end_position, move_object"
},
```
* *end_position* represent the target *[x, y]* coordinates of the robot arm end point of the movement around circular path. Note that only the arm moves while its base remains fixed.
* *move_object* is a boolean indicating whether the robot moves an object (*True*) or simply moves its arm without carrying an object (*False*).
* Robots without actions in a certain step should not be included.
* One robot can only be moved once in each step, which means that no repeated keys are allowed in the same step.
Ensure your output forms a valid JSON dictionary of next-step plan.



## Collision Avoidance Rules:
Your plan must strictly avoid collisions, as follows:
* Robot-Robot Collision
  * Each robot arm always swings along a smooth **circular** path around its base.
  * Two robot arms cannot occupy the same position at the end of a move.  
  * Their curved paths must not cross or share any point during the move.  
  * Sometimes a robot needs to move its arm to a safe position to avoid collision between another robot that move its arm to reach an object.
  * **Example:**  
    * robot_0 swings from [0.25, 0.25] to [0.75, 0.75] and robot_1 swings from [0.25, 0.75] to [0.75, 0.25] at the same time. Both arcs pass through [0.5, 0.5], causing a collision.
* Object-Object Collision
  * Two objects cannot occupy the same (x, y) at any time.  
  * If you move more than one object at once, they must have different drop-off points and non-crossing straight-line paths.
* Robot-Object Collision
  * An arm's circular path must not sweep through any object it isn't carrying.  
  * Before moving, confirm the curved trajectory does not pass over another object's position.


## Plan Efficiency Considerations:
* The exeuction time of your plan involves simultaneous robot arm movements from their current positions to specified target positions.
* Each robot arm moves at a constant speed of 0.5 units/time.
* The duration of the plan is determined by the longest single-arm movement within it.
* You should aim to minimize the execution time while ensuring collision-free movements and successful object placements.


## Example Input & Output:
Input:
<observation>
Object positions:
    Object 1: [0.55, 1.65]
Target positions:
    Object 1 target: [1.65, 0.55]
Robot positions:
    Robot 1: base [1.1, 1.1], arm [1.24, 0.61]
    Robot 2: base [1.1, 2.2], arm [1.24, 1.71]
</observation>

* Output:
<think> Let's understand the scenerio ... </think>
```json
{
  "Robot 1": "Move [0.55, 1.65] False"
}
```

Now work on the following problem given by user:

<observation>
{mapstate}
</observation>
\end{fancyminted}

\subsection{Prompt for Reasoning Behavior Probing}\label{subsec:checknumber-prompt}

We list the prompt for the reachability check in List~\ref{lst:reachcheck}, and the prompt for the collision check in List~\ref{lst:collisioncheck}.

\begin{fancyminted}{markdown}{Prompt for GPT-4o to count reachability check}{lst:reachcheck}
How many reachability checks about the robot's movement are presented in the following reasoning trace? For example, a sentence like 'Robot 0 (base [1.0, 1.0]) can reach [0.25, 0.25]' counts as one verification. Give me an integer number without saying anything else.
The reasoning trace is: 
{trace}
\end{fancyminted}

\begin{fancyminted}{markdown}{Prompt for GPT-4o to count collision check}{lst:collisioncheck}
How many collision checks about the robot's movement are presented in the following reasoning trace? For example, a sentence like 'Robot 0 moves to [0.25, 0.25], and Robot 1 moves to [0.25, 0.25]. They may collide with each other.' counts as one verification. Give me an integer number without saying anything else.
The reasoning trace is: 
{trace}
\end{fancyminted}

\end{document}